\documentclass[journal,twoside,web]{ieeecolor}
\usepackage{generic}
\usepackage{cite}
\usepackage{amsmath,amssymb,amsfonts,mathtools}
\usepackage{algorithm,algorithmic}
\usepackage{graphicx}
\usepackage{textcomp}
\usepackage{url}
\usepackage{balance}
\usepackage{booktabs}
\usepackage{array}
\usepackage{comment}

\makeatletter
\let\NAT@parse\undefined
\makeatother
\usepackage{hyperref}
\graphicspath{{figures/}}
\allowdisplaybreaks
\def\BibTeX{{\rm B\kern-.05em{\sc i\kern-.025em b}\kern-.08em
    T\kern-.1667em\lower.7ex\hbox{E}\kern-.125emX}}

\newcommand{\R}{\mathbb{R}}

\newcommand{\E}{\mathbb{E}}

\newcommand{\cS}{\mathcal{S}}

\newcommand{\cA}{\mathcal{A}}

\newcommand{\cP}{\mathcal{P}}

\newcommand{\Proj}{\mathrm{Proj}}

\newcommand{\norm}[1]{\left\lVert #1 \right\rVert}

\newcommand{\Dnorm}[1]{\left\lVert #1 \right\rVert_D}

\newcommand{\pos}[1]{\left[#1\right]_+}

\newtheorem{theorem}{Theorem}

\newtheorem{lemma}[theorem]{Lemma}

\newtheorem{proposition}[theorem]{Proposition}

\newtheorem{corollary}[theorem]{Corollary}

\newtheorem{assumption}{Assumption}

\newtheorem{remark}{Remark}

\newcommand{\cQ}{\mathcal Q}

\newcommand{\clip}{\operatorname{Clip}}
\newcommand{\argmax}{\operatorname*{arg\,max}}
\newcommand{\argmin}{\operatorname*{arg\,min}}

\newcommand{\Sym}{\mathbb S}
\newcommand{\inner}[2]{\left\langle #1,#2\right\rangle}

\begin{document}

\title{Finite-Time Convergence of Single-Trajectory Chi-Square Robust Q-Learning With Linear Function Approximation}

\author{Saptarshi Mandal, Yashaswini Murthy, and R. Srikant
\thanks{ This document is a preprint.}%
\thanks{S. Mandal is a PhD candidate at ECE, UIUC (email: smandal4@illinois.edu); Y. Murthy is an assistant professor at ORIE, UT Austin (email: yashaswini.murthy@austin.utexas.edu); R. Srikant is a professor at ECE, CSL, and the director of NCSA at UIUC (email: rsrikant@illinois.edu).}%
}
\date{}

\maketitle
\thispagestyle{empty}
\begin{abstract}
Distributionally robust reinforcement learning seeks policies that remain effective when the deployment environment differs from the one that generated the training data. We study model-free robust Q-learning with $\chi^2$ uncertainty sets and linear function approximation, using data from a single trajectory of an unknown nominal MDP. Evaluating the $\chi^2$ robust Bellman target introduces the square
root of a conditional second moment, which cannot be estimated
unbiasedly from one transition, while the projected robust Bellman
operator need not be contractive. We address these obstacles through a variational reformulation of the robust Bellman target and a blockwise frozen-target scheme, and establish a finite-time error bound relative to the optimal robust Q-function for every $\gamma\in(0,1)$. A neural-network experiment illustrates how the variational target can
be used in a continuous-state nonlinear-control task.
\end{abstract}


\section{Introduction}
\label{sec:introduction}

Reinforcement-learning algorithms are often trained under a nominal transition model and then deployed in an environment whose dynamics need not coincide with that model due to simulator mismatch, unmodeled dynamics, or incomplete system identification.  Distributionally robust reinforcement learning (DRRL) addresses this mismatch by optimizing performance against the worst transition kernel in a prescribed neighborhood of the nominal transition kernel.

We study model-free robust Q-learning when data are generated by an
unknown nominal transition kernel $P_0$ under a behavior policy $\pi_b$.
The learner does not know $P_0$, has no generative model, and cannot
sample arbitrary state--action pairs. The goal is to learn the optimal
robust Q-function---and hence a robust greedy policy---under an
$(s,a)$-rectangular $\chi^2$ uncertainty set
\cite{iyengar2005robust}. Because the state--action space may be large,
we approximate the robust Q-function linearly as
$Q(s,a)\approx\phi(s,a)^\top\theta$.

We focus on $\chi^2$ uncertainty because it models deployment shifts
that reweight successor outcomes already present under the nominal
dynamics. Such shifts arise, for example, when the frequencies of
modeled disturbance or actuator-failure modes change without introducing
new transition mechanisms. Because the $\chi^2$ divergence measures
likelihood-ratio deviations relative to the nominal law, it yields a
nominal-dependent, variance-sensitive geometry that differs from total
variation and $R$-contamination
\cite{duchi2019variance,shi2025curiouspricedistributionalrobustness,
panaganti2022sample}.

The exact robust Bellman operator is a $\gamma$-contraction in
$\ell_\infty$ under $(s,a)$-rectangular uncertainty
\cite{iyengar2005robust,nilim2005robust}. Learning its fixed point from
one trajectory nevertheless creates two difficulties.

The first is statistical. The inner $\chi^2$-robust expectation is
commonly evaluated through a scalar dual whose objective contains the
square root of a nominal conditional second moment. One observed
successor gives an unbiased estimate of that moment, but not of its
square root; differentiating the dual instead produces a ratio of
conditional moments whose denominator can vanish. Liang et al.\
\cite{liang2024single} address this issue in the tabular setting by
tracking the moments on faster time scales and updating Q-values on a
slower time scale. Their convergence analysis is asymptotic, while
their function-approximation implementation is empirical.

The second difficulty arises from function approximation. Projection
onto a linear feature class need not preserve the contraction of the
exact robust Bellman operator for general $\gamma\in(0,1)$. Existing
analyses therefore impose projected-contraction conditions that couple
the discount factor to the ambiguity model or assume additional
transition structure
\cite{zhou2023natural,ma2022distributionally,tang2024robust}.

We address the two difficulties separately. First, we apply a
variational identity to the scalar dual before stochastic
approximation. The resulting positive scale variable makes the
conditional moment enter the objective linearly. We represent the dual and scale functions using feature-based
parameters shared across state--action pairs and learn them jointly
from individual transitions. Because the resulting gradients
contain inverse powers of the scale, we impose a positive floor and
quantify its induced bias. Second, we freeze the value target within
each outer block. This separates within-block estimation and function
approximation from across-block Bellman iteration, allowing the latter
to use contraction of the exact robust operator rather than of its
projection.

\subsection{Contributions}
\label{sec:contributions}

The paper makes two main contributions.

\textbf{1) A finite-time Q-learning guarantee from one trajectory.}
We establish a finite-time $\ell_\infty$ error bound relative to the
optimal robust Q-function for discounted, model-free Q-learning with
state--action-rectangular $\chi^2$ uncertainty, fixed linear
Q-function approximation, and data from one continuing Markovian
behavior trajectory. The bound separates errors that decrease with the two within-block
sample sizes from approximation errors of the Q and auxiliary function
classes and the explicit bias introduced by imposing a positive lower
bound on the scale variable. It holds for every $\gamma\in(0,1)$ under geometric-mixing and
Q-feature-conditioning assumptions, without generative access or
contraction of the projected robust Bellman operator. When the selected
function classes support a prescribed accuracy $\epsilon$, the
resulting sufficient trajectory length is
$\widetilde O(\epsilon^{-6})$ for fixed problem and function-class
parameters.

We obtain this result using blockwise frozen targets. Within each
block, the algorithm estimates a projected robust Bellman update for a
fixed target; across blocks, the analysis invokes contraction of the
exact, unprojected robust Bellman operator. This separates function approximation from Bellman iteration and
adapts ideas from fitted value iteration and target-network Q-learning
\cite{munos2008finite,chen2023target} to robust optimality with jointly
estimated auxiliary functions. To our knowledge, existing guarantees do
not simultaneously cover standard $\chi^2$ uncertainty, linear
function approximation, robust control, one continuing trajectory, and
every $\gamma\in(0,1)$ without assuming contraction of a projected robust Bellman operator.

\textbf{2) Variational learning of the $\chi^2$ robust target with
shared function approximation.}
The scalar dual of the $\chi^2$-robust expectation contains the square
root of a conditional second moment. We apply the identity
$\sqrt z=\inf_{u>0}\{z/(2u)+u/2\}$ for $z \geq 0$ at the objective level, before
sampling. This makes the conditional second moment enter the objective linearly,
so one observed transition provides an unbiased estimate of the
conditional objective and conditionally unbiased gradients for the
visited state--action pair.

Rather than solving a separate dual problem at every state--action
pair, we represent both the state--action-dependent dual maximizer and
the new scale variable by shared, linear-in-parameter function classes.
The resulting objective is jointly concave, and the dual and scale parameter blocks can be learned jointly on one time
scale. We also impose a positive floor \(u\geq\ell\) on the scale variable $u$ to
bound the inverse-scale terms in the sample gradients, and show that
the resulting bias is \(O(\ell)\). The
analysis explicitly quantifies the approximation errors of both
auxiliary function classes rather than assuming that their pointwise
optimizers are represented exactly.

We complement the theory with two experiments. A tabular experiment
evaluated against a high-accuracy robust dynamic-programming reference
illustrates convergence toward the robust solution. A continuous-state
Pendulum experiment studies a practical neural adaptation outside the
finite-time theory.

\subsection{Related Work}
\label{sec:related}
\textbf{Robust MDPs and tabular DRRL:}
Robust MDPs with rectangular uncertainty sets were developed in the
classical dynamic-programming literature, e.g.,
\cite{iyengar2005robust,nilim2005robust}. Recent tabular DRRL results obtain
finite-sample guarantees through model estimation, robust value iteration,
offline coverage assumptions, or generative access, e.g.,
\cite{xu2023improved,shi2024distributionally}. Simulator-based model-free work includes KL-specific multilevel Monte
Carlo (MLMC) \cite{liu2022distributionally,wang2023finite},
KL-specific minibatch and variance-reduced methods
\cite{wang2024variance}, and thresholded MLMC for TV, KL, and
\(\chi^2\) uncertainty \cite{wang2024model}; these results are tabular and rely on the ability to resample selected state--action pairs. The
closest result under trajectory data is Liang et al.
\cite{liang2024single}, which treats the Cressie--Read family using
three-timescale moment tracking and gives an asymptotic tabular analysis; its
function-approximation implementation is empirical.

\textbf{Robust RL with function approximation:}
Existing function-approximation guarantees cover only parts of our setting.
Wang and Zou \cite{wang2021online} analyze robust TDC from one trajectory
under $R$-contamination, but their function-approximation guarantee concerns
stationarity of a projected policy-evaluation objective rather than the
optimal robust Q-function. The projected-Bellman analyses in
\cite{tamar2014scaling,roy2017reinforcement,badrinath2021robust} establish
convergence under pointwise transition-dominance or related discount-factor
conditions. 

Zhou et al.\ \cite{zhou2023natural} observe that several earlier
function-approximation methods, e.g., \cite{tamar2014scaling,roy2017reinforcement,badrinath2021robust} explicitly or implicitly assume an
oracle for approximately evaluating the robust Bellman operator. They address this issue and
obtain finite-time robust-policy guarantees using double-sampling and
feature-dependent integral probability metric (IPM) ambiguity sets whose targets admit unbiased
estimation from nominal samples under simulator access. Their setting
is complementary to ours: double sampling uses multiple successors for
selected state--action pairs, whereas their IPM ambiguity set differs
from the standard nominal-dependent $\chi^2$ ball considered here.

For the IPM construction, a sufficient feature-dependent
ambiguity-radius condition for projected contraction is proportional to
$(1-\gamma)/\gamma$. Their constant-stepsize policy guarantee also
invokes the robust optimality-gap bound in Lemma~9, which is stated
under the pointwise transition-dominance Assumption~2
\cite[Thm.~9, Lem.~9, and Prop.~3]{zhou2023natural}; the authors note that this
assumption cannot be guaranteed uniformly for canonical
$f$-divergence balls. Thus, existing function-approximation results do
not jointly cover standard $\chi^2$ uncertainty, one continuing
trajectory, and robust-control convergence without projected
contraction.


\textbf{Complementary assumptions or learning regimes:} Other function-approximation guarantees use complementary assumptions or
learning regimes. These include fitted Q-iteration for regularized MDPs
under robust exploration or coverage assumptions in
 \cite{panaganti2024model}, robust offline RL under partial coverage
in \cite{ma2022distributionally}, methods exploiting linear
or otherwise structured transition models in
 \cite{tang2024robust} and \cite{pmlr-v238-liu24d}, online
finite-horizon work with active exploration, e.g., 
\cite{ghosh2026orvit,ghosh2025scaling} and average-reward criterion in 
\cite{xu2025finite,roch2025finite,chen2025sample}. These regimes differ from the
discounted, off-policy, continuing-trajectory problem studied here.

Table~\ref{tab:comparison} summarizes the scope of the closest
theoretical guarantees.

\begin{table*}[t]
\centering
\caption{Closest theoretical guarantees. ``Single trajectory'' denotes
data from a fixed behavior chain without selected (state-action)-pair resampling;
\(Q_\chi^*\) denotes the optimal robust Q-function.}\label{tab:comparison}
\footnotesize
\setlength{\tabcolsep}{2.6pt}
\renewcommand{\arraystretch}{1.05}
\begin{tabular}{
@{}>{\raggedright\arraybackslash}p{0.115\textwidth}
>{\raggedright\arraybackslash}p{0.2\textwidth}
>{\raggedright\arraybackslash}p{0.145\textwidth}
>{\raggedright\arraybackslash}p{0.195\textwidth}
>{\raggedright\arraybackslash}p{0.3\textwidth}@{}}
\toprule
Work &
Ambiguity/method &
Data access &
Function approximation theory &
Type of guarantee \\
\midrule
Wang and Zou \cite{wang2021online}
&
\(R\)-contamination; robust TDC
&
\textbf{Single trajectory}
&
Linear critic for policy evaluation
&
Convergence to a stationary point of a projected policy-evaluation
objective
\\
Liang et al. \cite{liang2024single}
&
Cressie--Read family (including $\chi^2$ set); moment tracking with three-timescale SA
&
\textbf{Single trajectory}
&
Tabular theory; neural function approximation is empirical
&
Asymptotic convergence to the optimal robust Q-function under stated
regularity conditions on the multi-timescale drift maps
\\
Wang et al. \cite{wang2024model}
&
TV, KL, and \(\chi^2\); thresholded multilevel MC
&
Generative model
&
Tabular
&
Finite-sample and sample-complexity guarantees for tabular robust
Q-learning
\\
Zhou et al. \cite{zhou2023natural}
&
Double-Sampling (DS) or IPM uncertainty; natural actor--critic
&
\textbf{On-policy nominal trajectories}; DS additionally draws multiple
successors per visited pair
&
\textbf{Linear theory for DS/IPM; general-function theory for DS}
&
Finite-time robust-policy bound under projected-contraction and
transition-dominance conditions
\\
\midrule
\textbf{This paper}
&
\(\boldsymbol{\chi^2}\); joint dual/scale SA
with blockwise Q updates
&
\textbf{Single continuing Markovian trajectory}
&
\textbf{Linear function approximation}
&
\textbf{Finite-time \(\ell_\infty\) bound relative to \(Q_\chi^*\), for all \(\gamma\in(0,1)\), up to approximation and floor errors}
\\
\bottomrule
\end{tabular}
\end{table*}

\section{Problem Formulation and Robust Bellman Operator}
\label{sec:model}

We consider a finite-state, finite-action, infinite-horizon discounted
MDP $(\cS,\cA,P_0,r,\gamma)$. For $i=(s,a)$, the deterministic reward
satisfies $r(i)\in[-1,1]$ and is observed whenever $i$ is visited. The
learner knows neither the reward table nor $P_0$.

Data arrive along one continuing trajectory generated by a fixed
behavior policy $\pi_b$. At time $k$, the learner observes
$r(S_k,A_k)$ and
\[
S_{k+1}\sim P_0(\cdot\mid S_k,A_k),
\qquad
A_{k+1}\sim\pi_b(\cdot\mid S_{k+1}).
\]
The process cannot be reset, and additional successors cannot be
requested from a selected state--action pair.

\subsection{\texorpdfstring{Rectangular $\chi^2$ Uncertainty}
{Rectangular Chi-Square Uncertainty}}

For each state--action pair $i=(s,a)$, define the $\chi^2$ uncertainty
set of radius $\delta>0$ as
\begin{equation}
\label{eq:chi-set}
\cP_i^\chi
:=
\left\{
q\in\Delta_{\cS}:\;
q\ll P_0(\cdot\mid i),\;
D_\chi(q\|P_0(\cdot\mid i))\le \delta
\right\},
\end{equation}
where, for probability vectors \(q,p\in\Delta_{\mathcal S}\) satisfying
\(q\ll p\),
\begin{equation}
D_\chi(q\|p)
:=
\sum_{s'\in\operatorname{supp}(p)}
\frac{(q(s')-p(s'))^2}{p(s')}.
\label{eq:chi-divergence}
\end{equation}

We adopt the standard $(s,a)$-rectangular uncertainty model studied in 
\cite{iyengar2005robust,nilim2005robust}, under which the uncertainty set is defined as
\begin{equation}
    \label{eq:uncertainty}
    \cP^\chi
=
\bigotimes_{i\in\cS\times\cA}\cP_i^\chi .
\end{equation}

For a value vector $V$, define the worst-case conditional expectation
\begin{equation}
\sigma_i^\chi(V)
:=
\inf_{q\in\cP_i^\chi}q^\top V.
\label{eq:chi_DRO}
\end{equation}
For a bounded Q-function $Q$, let
$V_Q(s)=\max_{a\in\cA}Q(s,a)$ and define
\begin{equation}
(\mathcal T_\chi Q)(i)
=
r(i)+\gamma\sigma_i^\chi(V_Q).
\label{eq:robust-bellman}
\end{equation}
Under rectangularity, $\mathcal T_\chi$ is a
$\gamma$-contraction in $\ell_\infty$
\cite{iyengar2005robust,nilim2005robust}. Standard robust-MDP results
identify its unique fixed point with the optimal robust Q-function,
denoted $Q_\chi^*$. A policy greedy with respect to $Q_\chi^*$ is
robust optimal. Our objective is therefore to estimate this fixed point
from the single nominal trajectory using the prescribed function
classes.

Let $c_\delta=\sqrt{1+\delta}$.   The robust expectation in
\eqref{eq:chi_DRO} admits the scalar dual representation
\cite{duchi2019variance,duchi2021learning}
\begin{align}
\sigma_i^\chi(V)
& = \sup_{\eta\in\R}F_V^\chi(i;\eta) ,
\label{eq:chi-dual}\\
F_V^\chi(i;\eta)
&=\eta-c_\delta\sqrt{Z_V(i;\eta)},
\label{eq:scalar-dual-objective}\\
Z_V(i;\eta)
&=\E_{S'\sim P_0(\cdot\mid i)}[\pos{\eta-V(S')}^2\mid i] \nonumber.
\end{align}

Thus, the distributional minimization reduces to a scalar
maximization. However, the square root in
\eqref{eq:scalar-dual-objective} prevents an unbiased one-transition
plug-in estimate; the next section removes this obstruction at the
objective level.

\section{A Variational Reformulation and Function Approximation}
\label{sec:variational}

The obstacle in \eqref{eq:scalar-dual-objective} is not the estimation
of the conditional second moment itself: for fixed $(i,\eta)$, one
successor state gives the unbiased sample
\[
\bigl(\eta-V(S')\bigr)_+^2
\]
of $Z_V(i;\eta)$. The difficulty comes from applying the square root
before stochastic approximation. We therefore remove this nonlinearity
at the objective level.

For $z\geq0$, define \[q_z(u)=z/(2u)+u/2\] for $u>0$. Then
\begin{equation}
\sqrt z=\inf_{u>0}q_z(u),
\qquad
\argmin_{u>0}q_z(u)=\{\sqrt z\}
\quad (z>0).
\label{eq:qol}
\end{equation}
Applying this identity to \eqref{eq:chi-dual} gives
\begin{equation}
\sigma_i^\chi(V)
=
\sup_{\eta\in\R,\;u>0}
\left\{
\eta-\frac{c_\delta}{2}
\left(
\frac{Z_V(i;\eta)}{u}+u
\right)
\right\}.
\label{eq:variational-dual}
\end{equation}
For fixed $(\eta,u)$, the conditional moment now enters the objective
linearly. Substituting a single successor sample therefore gives an
unbiased estimate of the conditional objective. After the dual and
scale variables are parameterized, Proposition~\ref{prop:concavity}
shows that the same construction yields conditionally unbiased sample
gradients for the currently visited state--action pair.

\subsection{Scale Floor and Its Bias}

The auxiliary variable $u$ has a direct interpretation. When
$Z_V(i;\eta)>0$, the unconstrained minimizing value in
\eqref{eq:qol} is
\[
u=\sqrt{Z_V(i;\eta)},
\]
the conditional root-mean-square magnitude of the positive-part
residual. When $Z_V(i;\eta)=0$, the infimum is approached as
$u\downarrow0$.

Although the variational formulation removes the square root of the
conditional moment, its gradients contain the inverse factors $1/u$
and $1/u^2$. Allowing $u$ to approach zero would therefore make the
sample gradients unstable. We instead impose the positive floor
\[
u\geq\ell,
\qquad \ell\in(0,1].
\]
The next lemma shows that this stabilization changes the robust
surrogate by at most $c_\delta\ell/2$, uniformly over the target and
state--action pair.

\begin{lemma}[Scale-floor bias]
\label{lem:floor-bias}
For every \(s\ge0\) and \(\ell>0\), the minimizer of
\(q_{s^2}(u)=s^2/(2u)+u/2\) over \(u\in[\ell,\infty)\) is
\[
m_\ell(s):=\max\{\ell,s\}.
\]
Moreover,
\begin{equation}
q_{s^2}(m_\ell(s))-s
=
\frac{(\ell-s)_+^2}{2\ell}
\le \frac{\ell}{2}.
\label{eq:floor-bias}
\end{equation}
\end{lemma}

\begin{proof}
For \(u>0\),
\[
q_{s^2}'(u)
=
\frac12-\frac{s^2}{2u^2},
\quad
q_{s^2}''(u)
=
\frac{s^2}{u^3}\ge0.
\]
If \(s\ge\ell\), the unconstrained minimizer \(u=s\) is feasible and
the excess is zero. If \(0\le s<\ell\), then \(q_{s^2}\) is increasing
on \([\ell,\infty)\), so the constrained minimizer is \(u=\ell\), and
\(
q_{s^2}(\ell)-s
=
\frac{s^2}{2\ell}+\frac{\ell}{2}-s
=
\frac{(\ell-s)^2}{2\ell}
\le \frac{\ell}{2}.
\)
\end{proof}

For later reference, define the exact floor-constrained value
\begin{equation}
\sigma_{i,\ell}^{\chi}(V)
:=
\sup_{\eta\in\mathbb R,\;u\ge\ell}
\left\{
\eta-\frac{c_\delta}{2}
\left(
\frac{Z_V(i;\eta)}{u}+u
\right)
\right\}.
\label{eq:floor-constrained-value}
\end{equation}
Lemma~\ref{lem:floor-bias} and the scalar dual imply
\begin{equation}
0
\le
\sigma_i^\chi(V)-\sigma_{i,\ell}^\chi(V)
\le
\frac{c_\delta\ell}{2}.
\label{eq:robust-floor-bias}
\end{equation}

\subsection{Data and Behavior-Chain Assumptions}

The fixed behavior policy $\pi_b$ and nominal kernel $P_0$ generate the
state--action Markov chain
\[
I_k=(S_k,A_k).
\]
We impose the following standard geometric-mixing condition
\cite{srikant2019finite,li2022first,chen2023target,zhou2023natural}.

\begin{assumption}[Behavior-chain mixing]
\label{ass:mixing}
The chain $I_k$ is irreducible and aperiodic, with stationary
distribution $d$. There exist constants $C_{\rm mix}<\infty$ and
$\rho_{\rm mix}\in(0,1)$ such that
\begin{equation}
\sup_{i\in\mathcal S\times\mathcal A}
\norm{P_b^L(i,\cdot)-d}_{\rm TV}
\leq C_{\rm mix}\rho_{\rm mix}^L,
\qquad L\geq0,
\label{eq:mixing}
\end{equation}
where $P_b$ is the transition kernel of the state--action chain.
\end{assumption}

Let $D=\operatorname{diag}(d)$. For a vector indexed by
state--action pairs, write
\[
\norm{x}_D=(x^\top D x)^{1/2}.
\]
The constants $C_{\rm mix}$ and $\rho_{\rm mix}$ depend on the
behavior policy and nominal kernel.

\subsection{Approximation Classes}

We use fixed features to represent the three state--action-dependent
quantities learned by the algorithm: the robust Q-function, the scalar
dual function, and the positive scale function.

For the Q-function, let $\phi(i)\in\mathbb R^{d_\theta}$ and define
\[
Q_\theta(i)=\phi(i)^\top\theta,
\qquad
\theta\in\mathbb R^{d_\theta}.
\]
Let $\Phi$ be the matrix whose $i$-th row is $\phi(i)^\top$.

For the scalar dual function, let
$\psi(i)\in\mathbb R^{d_\nu}$ and define
\[
\eta_\nu(i)=\psi(i)^\top\nu,
\qquad
\nu\in\mathbb R^{d_\nu}.
\]
Let $\Psi$ be the matrix whose $i$-th row is $\psi(i)^\top$.

For fixed $V$ and $\nu$, the optimal floor-constrained scale at
state--action pair $i$ is
\begin{equation}
u^*_{\ell,V,\nu}(i)
=
\max\left\{
\ell,
\sqrt{
\E\!\left[
\bigl(\psi(i)^\top\nu-V(S')\bigr)_+^2
\mid i
\right]}
\right\}.
\label{eq:optimal-scale-map}
\end{equation}
Thus, the scale class must represent a nonnegative function while
retaining affine dependence on its learnable parameter, as required
for joint concavity. For the theoretical results, let
$\chi(i)\in\mathbb R^{d_\chi}$ be a signed feature vector and define
\begin{equation}
\begin{aligned}
X_i&:=\chi(i)\chi(i)^\top,\\
g_\Theta(i)
&:=\inner{\Theta}{X_i}_F
=\chi(i)^\top\Theta\chi(i),\\
u_{\Theta,\ell}(i)
&:=\ell+g_\Theta(i),
\end{aligned}
\label{eq:function-classes-aux}
\end{equation}
where $\Theta\in\mathbb S_+^{d_\chi}$ is a learnable
positive-semidefinite matrix. This representation permits signed
features while guaranteeing $g_\Theta(i)\geq0$, and it remains affine
in $\Theta$.

\begin{assumption}[Feature conditioning]
\label{ass:features}
For some $\mu_\phi \in (0,1)$,
\begin{equation}
\norm{\phi(i)}_2,\norm{\psi(i)}_2,\norm{\chi(i)}_2\leq1,
\qquad
\Phi^\top D\Phi\succeq\mu_\phi I.
\label{eq:feature-assumption}
\end{equation}
\end{assumption}

Only the Q-feature Gram matrix requires a positive lower bound.
Stage~1 is analyzed through its objective gap and does not require
analogous covariance lower bounds for the dual or scale features.

\subsection{Concavity of the Variational Objective}

Given one observed transition $(i,S')$ and a frozen value target $V$, define
\[
x_{V,\nu}(i,S'):=\bigl(\psi(i)^\top\nu-V(S')\bigr)_+.
\]
The one-transition variational objective is
\begin{equation}
\widehat\varphi_{V,\ell}(\nu,\Theta;i,S')
=
\psi(i)^\top\nu
-
\frac{c_\delta}{2}
\left[
\frac{x_{V,\nu}(i,S')^2}{u_{\Theta,\ell}(i)}
+
u_{\Theta,\ell}(i)
\right].
\label{eq:sample-objective}
\end{equation}

For a fixed state--action pair $i$, define the corresponding conditional
population objective by
\[
\bar\varphi_{V,\ell}(\nu,\Theta;i)
:=
\E_{S'\sim P_0(\cdot\mid i)}
\left[
\widehat\varphi_{V,\ell}(\nu,\Theta;i,S')
\right].
\]
The next proposition establishes the two properties needed in
Stage~1: joint concavity and unbiased one-transition gradients.

\begin{proposition}[Joint concavity and sample gradients]
\label{prop:concavity}
The sample objective in \eqref{eq:sample-objective} is jointly concave
and continuously differentiable in \((\nu,\Theta)\) on
\(\mathbb R^{d_\nu}\times\mathbb S_+^{d_\chi}\).
Abbreviate
\[
x=\bigl(\psi(i)^\top\nu-V(S')\bigr)_+,
\qquad
u=\ell+\inner{\Theta}{X_i}_F.
\]
Then its gradients are
\begin{equation}
\nabla_\nu\widehat\varphi_{V,\ell}
=
\left(1-c_\delta\frac{x}{u}\right)\psi(i),
\quad
\nabla_\Theta\widehat\varphi_{V,\ell}
=
\frac{c_\delta}{2}
\left(\frac{x^2}{u^2}-1\right)X_i.
\label{eq:sample-gradients}
\end{equation}

Moreover, the stacked gradient satisfies
\begin{equation}
\mathbb E_{S'\sim P_0(\cdot\mid i)}
\left[
\nabla\widehat\varphi_{V,\ell}
(\nu,\Theta;i,S')
\right]
=
\nabla\bar\varphi_{V,\ell}
(\nu,\Theta;i).
\label{eq:visited-index-unbiasedness}
\end{equation}
Thus, conditional on the visited state--action pair, one transition
provides an unbiased gradient of the corresponding conditional
population objective.

\end{proposition}

\begin{proof}
\((a,u)\longmapsto \frac{(a_+)^2}{u},
\quad u>0,\)
is jointly convex because
\[
\frac{(a_+)^2}{u}
=
\sup_{v\geq0}\{2va-v^2u\},
\]
which is a pointwise supremum of affine functions of $(a,u)$. Here,
$a=\psi(i)^\top\nu-V(S')$ and
$u=\ell+\inner{\Theta}{X_i}_F$ are affine in $(\nu,\Theta)$, and
$u\geq\ell>0$.
The function $a\mapsto(a_+)^2$ is continuously differentiable, including at
$a=0$. Direct differentiation gives \eqref{eq:sample-gradients}. Conditional unbiasedness follows by taking expectation over
\(S'\sim P_0(\cdot\mid i)\). Interchanging differentiation and
expectation is valid because the successor-state space is finite and
the denominator is bounded below by \(\ell>0\).
\end{proof}

\begin{remark}[Why the direct scalar-dual update is delicate]
\label{rem:direct-dual-nonsmoothness}
The moment-tracking method of Liang et al.\
\cite{liang2024single} operates directly on the scalar objective
$F_V^\chi(i;\eta)$ in \eqref{eq:scalar-dual-objective}. Whenever
$Z_V(i;\eta)>0$, its derivative is
\begin{equation}
\frac{\partial F_V^\chi(i;\eta)}{\partial\eta}
=
1-c_\delta
\frac{
\E[(\eta-V(S'))_+\mid i]
}{
\sqrt{\E[(\eta-V(S'))_+^2\mid i]}
}.
\label{eq:direct-dual-gradient}
\end{equation}
The objective need not be differentiable even on the natural bounded
range of the dual variable. To see this, let
\(
\underline v_i
:=
\min_{s'\in\operatorname{supp}(P_0(\cdot\mid i))}V(s'),
\quad
p_i^{\min}
:=
P_0\!\left(V(S')=\underline v_i\mid i\right)>0.
\)
For sufficiently small $h>0$,
\(
F_V^\chi(i;\underline v_i-h)=\underline v_i-h,
\qquad
F_V^\chi(i;\underline v_i+h)
=
\underline v_i+
\left(1-c_\delta\sqrt{p_i^{\min}}\right)h.
\)
The left and right derivatives at $\underline v_i$ are therefore
$1$ and $1-c_\delta\sqrt{p_i^{\min}}$, respectively. At the same
point, the ratio in \eqref{eq:direct-dual-gradient} has the
indeterminate form $0/0$. Consequently, without an additional
nondegeneracy condition or smoothing, global Lipschitz regularity of
the direct dual-update mean field is not automatic; such regularity is
invoked in \cite[Condition~B.9 and
Theorem~B.18]{liang2024single}.

The reformulation \eqref{eq:variational-dual} instead removes the
square root before stochastic approximation. Its only denominators
are powers of the positive scale $u$. Enforcing $u\geq\ell>0$, together
with clipped targets and compact parameter domains, yields uniformly
bounded sample gradients. Lemma~\ref{lem:floor-bias} quantifies the
corresponding $O(\ell)$ bias explicitly.  Thus, our analysis avoids imposing global Lipschitz regularity or a
nondegeneracy condition on the direct scalar-dual update, including in
the tabular one-hot specialization.
\end{remark}
\begin{remark}[Scale function class choice]
\label{rem:psd-motivation}
The positive semidefinite (PSD) representation permits signed scale features while enforcing
nonnegativity and retaining affine dependence on a convex parameter.
When entrywise nonnegative features are available, the same analysis
also permits a cheaper nonnegative vector parameterization.
Section~\ref{sec:computation} states that model explicitly and compares
its projection cost and expressivity with the PSD class.
\end{remark}

\subsection{Parameter Domains and Approximation Errors}

 Let \(B_Q=(1-\gamma)^{-1}\), and let \(\clip\) denote componentwise
clipping onto \([-B_Q,B_Q]\).
 Since \(|r(i)|\le1\), the robust fixed point satisfies
\begin{align}
\|Q_\chi^*\|_\infty
=
\|\mathcal T_\chi Q_\chi^*\|_\infty
\le
1+\gamma\|Q_\chi^*\|_\infty,
\end{align}
and therefore
\begin{equation}
\|Q_\chi^*\|_\infty\le B_Q,
\quad
\clip(Q_\chi^*)=Q_\chi^*.
\label{eq:optimal-Q-bound}
\end{equation}
 
Also, let
$\Pi_\phi$ be the $D$-orthogonal projection onto $\mathrm{range}(\Phi)$:
\begin{equation}
\cQ=\{\clip(\Phi\theta):\theta\in\R^{d_\theta}\},
\quad
\Pi_\phi f=\Phi(\Phi^\top D\Phi)^{-1}\Phi^\top Df.
\end{equation}
The Q-function approximation error is defined as 
\begin{equation}
\mathcal E_Q=
\sup_{Q\in\cQ}
\norm{\clip(\Pi_\phi\mathcal T_\chi Q)-\mathcal T_\chi Q}_\infty.
\label{eq:Q-approx}
\end{equation}

Thus, $\mathcal E_Q$ measures approximate closure of the Q-function
class under the clipped robust Bellman updates encountered by the
algorithm. Representability of $Q_\chi^*$ alone does not imply
$\mathcal E_Q=0$.

Before introducing the compact parameter domains, we record that
bounded value targets admit bounded scalar-dual maximizers. Thus, the
unbounded domain in \eqref{eq:chi-dual} is not intrinsic to the
optimization problem.

\begin{lemma}[Compact dual interval]
\label{lem:eta-bound}
Define
\begin{equation}
B_\eta(B)
:=
B\left(1+\frac{1}{\sqrt{\delta}}\right).
\label{eq:eta-bound}
\end{equation}
If \(\|V\|_\infty\le B\), then a maximizer of
\eqref{eq:chi-dual} exists in
\(
[-B,B_\eta(B)].
\)
\end{lemma}

\begin{proof}
For \(\eta\le\min_sV(s)\), \(F_V^\chi(i;\eta)=\eta\), so no point below
\(-B\) is optimal. The claim is immediate if \(B=0\). For \(B>0\) and
\(\eta\ge B\), let \(m_i=\E[V(S')\mid i]\) and
\(v_i=\operatorname{Var}(V(S')\mid i)\). Then
\(
F_V^\chi(i;\eta)
=\eta-c_\delta\sqrt{(\eta-m_i)^2+v_i},
\qquad v_i\le B^2.
\)
If \(\eta\ge B_\eta(B)\), then
\(\eta-m_i\ge B/\sqrt{\delta}\), so
\(v_i\le\delta(\eta-m_i)^2\). Since \(c_\delta^2=1+\delta\),
this implies
\(
\partial_\eta F_V^\chi(i;\eta)
=
1-c_\delta\frac{\eta-m_i}
{\sqrt{(\eta-m_i)^2+v_i}}
\le0.
\)
Thus $F_V^\chi(i;\eta)$ is nonincreasing above \(B_\eta(B)\), and continuity
on the remaining compact interval proves attainment.
\end{proof}

We optimize over the compact convex set
\begin{equation}
\begin{aligned}
\mathcal M_\nu
&=\{\nu:\norm{\nu}_2\leq R_\nu\},\\
\mathcal M_\Theta
&=\{\Theta\in\Sym^{d_\chi}:\Theta\succeq0,\,
\norm{\Theta}_F\leq R_\Theta\},\\
\mathcal W
&:=\mathcal M_\nu\times\mathcal M_\Theta,
\end{aligned}
\label{eq:parameter-sets}
\end{equation}

where $R_\nu,R_\Theta>0$. Equip \(\mathcal W\) with
\(\|(\nu,\Theta)\|^2=\|\nu\|_2^2+\|\Theta\|_F^2\); projection onto
\(\mathcal W\) separates across the two parameter blocks. For frozen \(V\) with \(\|V\|_\infty\le B_Q\), set
\(H:=R_\nu+B_Q\). Then \(0\le x\le H\), \(\|X_i\|_F\le1\), and
\begin{equation}
\ell\leq u_{\Theta,\ell}(i)\leq\ell+R_\Theta.
\label{eq:scale-range}
\end{equation}

Proposition~\ref{prop:concavity} and the preceding bounds imply
\begin{align}
&\norm{(\nabla_\nu\widehat\varphi_{V,\ell},\nabla_\Theta\widehat\varphi_{V,\ell})} \le G_\ell,
\label{eq:Gell}\\
&G_\ell=\left[
\left(1+\frac{c_\delta H}{\ell}\right)^2
+\frac{c_\delta^2}{4}\left(1+\frac{H^2}{\ell^2}\right)^2
\right]^{1/2}, \nonumber \\
&|\widehat\varphi_{V,\ell}|\le B_\varphi(\ell)
\coloneqq R_\nu+\frac{c_\delta}{2}
\left(\frac{H^2}{\ell}+\ell+R_\Theta\right).
\label{eq:Bphi}
\end{align}
For fixed problem parameters, $G_\ell=O(\ell^{-2})$ and $B_\varphi(\ell)=O(\ell^{-1})$.

For a frozen $V$, define the stationary population objective
\begin{equation}
J_{V,\ell}(\nu,\Theta)
=\sum_i d_i\E[\widehat\varphi_{V,\ell}(\nu,\Theta;i,S')\mid i]
\label{eq:population-objective}
\end{equation}
and $J_{V,\ell}^{\rm PSD,*}=\max_{(\nu,\Theta)\in\mathcal W}J_{V,\ell}(\nu,\Theta)$.

For fixed \(V\) and \(\nu\), define the unshifted root-mean-square
scale
\begin{equation}
s^{V,\nu}(i)
:=
\sqrt{Z_V(i;\psi(i)^\top\nu)}
\label{eq:unshifted-scale}
\end{equation}
and the stationary dual approximation loss
\begin{equation}
\mathcal A_\eta(V,\nu)
:=
\sum_i d_i
\left[
\sigma_i^\chi(V)
-(\psi(i)^\top\nu
-c_\delta s^{V,\nu}(i))
\right].
\label{eq:Aeta}
\end{equation}
The quantity \(\mathcal A_\eta(V,\nu)\) is nonnegative because
\(\psi(i)^\top\nu-c_\delta s^{V,\nu}(i)\) is a feasible value of the
scalar dual objective.

For each \(Q\in\cQ\), let
\begin{equation}
\mathcal N_Q^*
:=
\argmin_{\nu\in\mathcal M_\nu}
\mathcal A_\eta(V_Q,\nu).
\label{eq:best-dual-comparators}
\end{equation}
The dual and scale function-approximation errors are
\begin{align}
\mathcal E_\eta
&:=
\sup_{Q\in\cQ}
\min_{\nu\in\mathcal M_\nu}
\mathcal A_\eta(V_Q,\nu),
\label{eq:dual-fa-error}\\
\mathcal E_{\rm sc}
&:=
\sup_{Q\in\cQ}
\min_{\substack{\nu\in\mathcal N_Q^*\\
                 \Theta\in\mathcal M_\Theta}}
\Dnorm{g_\Theta-s^{V_Q,\nu}}.
\label{eq:scale-fa-error}
\end{align}

The inner minima are attained by compactness and continuity. Both
\(\mathcal E_\eta\) and \(\mathcal E_{\rm sc}\) may depend on the
problem geometry, ambiguity radius, approximation classes, and
parameter radii, but neither depends on the design floor \(\ell\).

The preceding definitions describe the primitive representation
errors. The next proposition shows how these errors, together with the
scale floor, determine the best variational objective attainable within
the chosen classes.

\begin{proposition}[Dual, scale, and floor decomposition]
\label{prop:fa-floor-decomposition}
For every \(\ell>0\),
\begin{align}
&\sup_{Q\in\cQ}
\left\{
\sum_i d_i \sigma_i^\chi(V_Q)
-J_{V_Q,\ell}^{\mathrm{PSD},*}
\right\}
\nonumber\\
&\quad\le
\underbrace{
\mathcal E_\eta
+\frac{c_\delta}{2}
\left(
\mathcal E_{\rm sc}
+\frac{\mathcal E_{\rm sc}^2}{\ell}
\right)
}_{\text{dual and scale function approximation}}
+
\underbrace{
\frac{c_\delta\ell}{2}
}_{\text{floor bias}}.
\label{eq:fa-floor-decomposition}
\end{align}
\end{proposition}

\begin{proof}
Fix \(Q\in\cQ\). By compactness and continuity, choose
\(\nu_Q\in\mathcal N_Q^*\) and
\(\Theta_Q\in\mathcal M_\Theta\) attaining the comparator in
\eqref{eq:scale-fa-error} for this \(Q\). Define
\[
s_i:=s^{V_Q,\nu_Q}(i),
\qquad
g_i:=g_{\Theta_Q}(i),
\qquad
e_i:=g_i-s_i.
\]
For arbitrary \(s,g\ge 0\), the definition of \(q_z\) gives
\begin{align}
q_{s^2}(\ell+g)-s
&=
\frac12
\left(
\frac{s^2}{\ell+g}+\ell+g
\right)-s =
\frac{(\ell+g-s)^2}{2(\ell+g)}
\nonumber\\
&\le
\frac{\ell}{2}
+\frac{|g-s|}{2}
+\frac{(g-s)^2}{2\ell}.
\label{eq:unshifted-scale-bound}
\end{align}

Indeed, writing $e=g-s$, the left-hand side is at most
$(\ell+e)/2$ when $e\geq0$, and at most
$(\ell+e)^2/(2\ell)$ when $e<0$; both are bounded by the
right-hand side of \eqref{eq:unshifted-scale-bound}.

Since
\[
J_{V_Q,\ell}^{\mathrm{PSD},*}
\ge
J_{V_Q,\ell}(\nu_Q,\Theta_Q),
\]
we obtain
\begin{align*}
&\sum_i d_i \sigma_i^\chi(V_Q)
-J_{V_Q,\ell}^{\mathrm{PSD},*} \le
\sum_i d_i \sigma_i^\chi(V_Q)
-J_{V_Q,\ell}(\nu_Q,\Theta_Q)\\
&\quad=
\mathcal A_\eta(V_Q,\nu_Q)
+c_\delta\sum_i d_i
\left[
q_{s_i^2}(\ell+g_i)-s_i
\right]\\
&\quad\le
\mathcal A_\eta(V_Q,\nu_Q)
+\frac{c_\delta}{2}
\left(
\ell+\sum_i d_i|e_i|
+\frac{1}{\ell}\sum_i d_i e_i^2
\right)\\
&\quad \underbrace{\le}_{(a)}
\mathcal A_\eta(V_Q,\nu_Q)
+\frac{c_\delta}{2}
\left[
\ell
+\left(\sum_i d_i e_i^2\right)^{1/2}
+\frac{1}{\ell}\sum_i d_i e_i^2
\right],
\end{align*}
where $(a)$ follows from Cauchy--Schwarz and
\(\sum_i d_i=1\). Using the definitions of
\(\mathcal E_\eta\), \(\mathcal E_{\rm sc}\), and $e_i$ we get
\eqref{eq:fa-floor-decomposition}.
\end{proof}

\section{Two-Stage Target-Network Algorithm}
\label{sec:algorithm}

Each outer block freezes
$\widehat Q_t=\clip(\Phi\widehat\theta_t)$ and
$V_t(s)=\max_a\widehat Q_t(s,a)$. Stage~1 uses $N$ transitions and
projected stochastic ascent to learn shared dual and scale functions
for this frozen target, returning the averaged parameters
$(\bar\nu_t,\bar\Theta_t)$. Stage~2 freezes these averages and uses
$K$ further transitions to regress the resulting robust Bellman
samples onto the Q-feature space with a decreasing stepsize. Its final
iterate defines the target for the next block. All stages and blocks
consume consecutive transitions from the single behavior trajectory;
the process is never reset.

\begin{algorithm}[t]
\caption{Variational $\chi^2$-robust Q-learning with a PSD scale
parameterization}
\label{alg:main}
\footnotesize
\begin{algorithmic}[1]
\STATE \textbf{Input:} integers \(T\ge0\) and \(N,K\ge1\), floor
\(\ell\in(0,1]\), stepsize parameters
\(\alpha_N,\beta_0,h_Q\), and behavior policy \(\pi_b\).
\STATE Initialize \(\widehat\theta_0\),
\(\widehat Q_0\gets\clip(\Phi\widehat\theta_0)\), and the trajectory
cursor \(n\gets0\).
\FOR{\(t=0,1,\ldots,T-1\)}
    \STATE Set \(\widehat Q_t=\clip(\Phi\widehat\theta_t)\) and
    \(V_t(s)=\max_{a\in\cA}\widehat Q_t(s,a)\).
    \STATE \textit{Stage 1: variational target estimation.} 
    \STATE Initialize
    \((\nu_{t,0},\Theta_{t,0})=(0,0)\).
    \FOR{\(k=0,1,\ldots,N-1\)}
        \STATE Read the next transition
        \((i,s')\gets(I_n,S_{n+1})\).
        \STATE Compute
        \(x_{t,k}=\pos{\psi(i)^\top\nu_{t,k}-V_t(s')}\) and
        \(u_{t,k}=\ell+\inner{\Theta_{t,k}}{X_i}_F\).
        \STATE \(g^\nu_{t,k}\gets
        \left(1-c_\delta\frac{x_{t,k}}{u_{t,k}}\right)\psi(i).
        \)
        \STATE \(g^\Theta_{t,k}\gets
        \frac{c_\delta}{2}
        \left(\frac{x_{t,k}^2}{u_{t,k}^2}-1\right)X_i.\)
        \STATE \(\nu_{t,k+1}\gets
        \Proj_{\mathcal M_\nu}
        \bigl(\nu_{t,k}+\alpha_Ng^\nu_{t,k}\bigr).\)
        \STATE \(\Theta_{t,k+1}\gets
        \Proj_{\mathcal M_\Theta}
        \bigl(\Theta_{t,k}+\alpha_Ng^\Theta_{t,k}\bigr).\)
        \STATE \(n\gets n+1\).
    \ENDFOR
    \STATE \(\displaystyle
    \bar\nu_t\gets N^{-1}\sum_{k=0}^{N-1}\nu_{t,k}\), \quad
    \(\displaystyle
    \bar\Theta_t\gets N^{-1}\sum_{k=0}^{N-1}\Theta_{t,k}\).
    \STATE \textit{Stage 2: linear Q regression.} 
    \STATE Initialize
\(\theta_{t,0}=0\).
    \FOR{\(m=0,1,\ldots,K-1\)}
        \STATE Read the next transition
        \((i,s')\gets(I_n,S_{n+1})\).
        \STATE Compute
        \(\widetilde x_{t,m}=
        \pos{\psi(i)^\top\bar\nu_t-V_t(s')}\) and
        \(\widetilde u_{t,m}=
        \ell+\inner{\bar\Theta_t}{X_i}_F\).
        \STATE \(\displaystyle
        \widehat f_{t,m}\gets\psi(i)^\top\bar\nu_t-
        \frac{c_\delta}{2}
        \left(
        \frac{\widetilde x_{t,m}^2}{\widetilde u_{t,m}}
        +\widetilde u_{t,m}
        \right).\)
        \STATE \(Y_{t,m}\gets r(i)+\gamma\widehat f_{t,m}\), \quad
        \(\beta_m\gets\beta_0/(m+h_Q)\).
        \STATE \(\displaystyle
        \theta_{t,m+1}\gets\theta_{t,m}+
        \beta_m
        \bigl[Y_{t,m}-\phi(i)^\top\theta_{t,m}\bigr]\phi(i).\)
        \STATE \(n\gets n+1\).
    \ENDFOR
    \STATE \(\widehat\theta_{t+1}\gets\theta_{t,K}\).
\ENDFOR
\STATE \textbf{Output:} \(\widehat Q_T=\clip(\Phi\widehat\theta_T)\).
\end{algorithmic}
\end{algorithm}

\section{Main Results}
\label{sec:main-results}

In this section, we present the main result of the paper. For \(\|V\|_\infty\le B_Q\) and
\((\nu,\Theta)\in\mathcal W\), define the mean Stage~2 target by

\begin{align}
\overline Y_{V,\nu,\Theta}(i)
&=
r(i)+\gamma f_{V,\ell}(i;\nu,\Theta),
\label{eq:mean-target}\\
f_{V,\ell}(i;\nu,\Theta)
&=
\psi(i)^\top\nu
-
\frac{c_\delta}{2}
\left[
\frac{Z_V(i;\psi(i)^\top\nu)}
     {\ell+g_\Theta(i)}
+\ell+g_\Theta(i)
\right].
\nonumber
\end{align}
Uniformly in these inputs, \eqref{eq:Bphi} gives
\begin{equation}
|Y_{t,m}|,
\quad
|\overline Y_{V,\nu,\Theta}(i)|
\leq B_Y(\ell),
\qquad
B_Y(\ell):=1+\gamma B_\varphi(\ell).
\label{eq:BY}
\end{equation}

We next define the finite-sample error envelopes appearing in the main
theorem. For $n\geq1$, let

\begin{equation}
\Lambda_n:=1+\log(n+1)+\log^2(n+1).
\label{eq:Lambda}
\end{equation}
Define
\begin{align}
\kappa_{\rm mix}
&:=
1+\frac{1+\log(C_{\rm mix}\vee1)}{\log(1/\rho_{\rm mix})},
\label{eq:kappa-mix}
\end{align}
and set
\begin{align}
&A_{\rm mix}:=1+2\kappa_{\rm mix},
&
D_{\rm mix}&:=4(1+\kappa_{\rm mix})  \nonumber \\
& R_{\mathcal W}
:=
\sqrt{R_\nu^2+R_\Theta^2}.
\label{eq:stage1-explicit-constants}
\end{align}
The Stage~1 error envelope is defined as
\begin{equation}
\overline\varepsilon_{\rm var}(N,\ell)
:=
A_{\rm mix}R_{\mathcal W}G_\ell
\sqrt{\frac{\Lambda_N}{N}}
+
D_{\rm mix}B_\varphi(\ell)\frac{\Lambda_N}{N}.
\label{eq:stage1-envelope}
\end{equation}

For the Q-stage, define
\begin{align}
p_Q&:=2\mu_\phi\beta_0,
&
\Lambda^Q_{K,h_Q}
&:=
1+\log(K+h_Q)
\label{eq:q-rate-definitions}
\end{align}
and
\begin{equation}
\mathfrak R_Q(K)
:=
\frac{\Lambda^Q_{K,h_Q}}{K+h_Q}.
\label{eq:q-rate}
\end{equation}
The constant \(C_Q\) given in
equation~\eqref{eq:CQ-explicit} is independent of \(N,K,\ell,T,\gamma,\delta\), and the frozen target after the factor
\(B_Y(\ell)\) has been extracted.  Write
\begin{equation}
\varepsilon_Q(K,\ell)
:=
B_Y(\ell)\sqrt{C_Q\mathfrak R_Q(K)}.
\label{eq:eps-Q}
\end{equation}
Finally, let
\begin{equation}
\Gamma_T:=\sum_{j=0}^{T-1}\gamma^j
=\frac{1-\gamma^T}{1-\gamma},
\qquad
\Gamma_0:=0.
\label{eq:GammaT}
\end{equation}

\begin{theorem}[Finite-Time Error Bound]
\label{thm:main}
Suppose that Assumptions~\ref{ass:features} and~\ref{ass:mixing} hold,
and let Algorithm~\ref{alg:main} use the parameter domains in
\eqref{eq:parameter-sets}. In
Algorithm~\ref{alg:main}, take \(\widehat\theta_0\) deterministic and use
\begin{equation}
\alpha_N
=
\frac{R_{\mathcal W}}{G_\ell\sqrt{N\Lambda_N}},
\qquad
\beta_m=\frac{\beta_0}{m+h_Q},
\label{eq:main-stepsizes}
\end{equation}
where \(\beta_0>0\), \(h_Q\ge\max\{1,2\beta_0\}\), and
\begin{equation}
p_Q=2\mu_\phi\beta_0>1.
\label{eq:q-clean-gain}
\end{equation}
Then, for all integers \(N,K\ge1\) and \(T\ge0\),
\begin{align}
&\E\norm{\widehat Q_T-Q_\chi^*}_\infty
\le{}
\gamma^T\norm{\widehat Q_0-Q_\chi^*}_\infty
\nonumber\\
&+\Gamma_T
\Bigg\{
\underbrace{\varepsilon_Q(K,\ell)}_{\text{Q-stage statistical error}}
+\frac{\gamma}{\mu_\phi}
\Bigg[
\underbrace{\overline\varepsilon_{\rm var}(N,\ell)}_{\text{variational-stage statistical error}}
\nonumber\\
&+
\underbrace{
\mathcal E_\eta
+\frac{c_\delta}{2}
\left(
\mathcal E_{\rm sc}
+\frac{\mathcal E_{\rm sc}^2}{\ell}
\right)
}_{\text{dual and scale function approximation error}}
+
\underbrace{
\frac{c_\delta\ell}{2}
}_{\text{floor bias}}
\Bigg]
+\mathcal E_Q
\Bigg\}.
\label{eq:main-bound}
\end{align}
\end{theorem}

The first term is the outer-loop transient. The braces contain the two
within-block statistical errors, the three function-approximation
errors, and the explicit floor bias; all accumulate through
\(\Gamma_T\leq(1-\gamma)^{-1}\).

\begin{remark}[Gain choice]
\label{rem:q-offset}
The condition $p_Q>1$ in \eqref{eq:q-clean-gain} is used to obtain the
clean $\widetilde O(K^{-1/2})$ Q-stage norm rate stated in the main
theorem. The recursion remains convergent for $p_Q>0$; the boundary case $p_Q=1$ and the slower regime
$0<p_Q<1$ are given in Remark~\ref{rem:q-general-gain}.
\end{remark}

The following sample complexity result is conditional on its displayed approximation
requirement; a fixed function class may therefore have a nonzero
accuracy floor. With \(\ell=\Theta(\epsilon)\), its statistical budgets
are \(N=\widetilde O(\epsilon^{-6})\) and
\(K=\widetilde O(\epsilon^{-4})\), so Stage~1 determines the exponent.

 Define
\begin{align}
\overline G
&:=
1+c_\delta H+\frac{c_\delta}{2}(1+H^2),
\nonumber\\
\overline B_\varphi
&:=
R_\nu+\frac{c_\delta}{2}(H^2+1+R_\Theta), \quad
\overline B_Y
:=
1+\gamma\overline B_\varphi.
\label{eq:problem-envelopes}
\end{align}

\begin{corollary}[Prescribed-accuracy sample complexity]
\label{cor:complexity}
Under the conditions of Theorem~\ref{thm:main}, fix
\(c_\ell\in(0,1]\) and set
\(\ell=c_\ell\epsilon\) for \(\epsilon\in(0,1)\).
Assume that the function-approximation and floor terms satisfy
\begin{align}
\underbrace{
\mathcal E_Q
+\frac{\gamma}{\mu_\phi}
\left[
\mathcal E_\eta
+\frac{c_\delta}{2}
\left(
\mathcal E_{\rm sc}
+\frac{\mathcal E_{\rm sc}^2}{c_\ell\epsilon}
\right)
\right]
}_{\text{function-approximation error}}
+
\underbrace{
\frac{\gamma c_\delta c_\ell}{2\mu_\phi}\epsilon
}_{\text{floor bias}}\le
\frac{(1-\gamma)\epsilon}{4}.
\label{eq:approximation-floor-condition}
\end{align}
Choose \(N,K\) so that
\begin{align*}
&\frac{N}{\Lambda_N}
 \ge 
\max\left\{
\left[
\frac{
8\gamma A_{\rm mix}R_{\mathcal W}\overline G
}{
\mu_\phi(1-\gamma)c_\ell^2
}
\right]^2\epsilon^{-6},
\frac{
8\gamma D_{\rm mix}\overline B_\varphi
}{
\mu_\phi(1-\gamma)c_\ell
}\epsilon^{-2}
\right\},\\
&\frac{K+h_Q}{\Lambda^Q_{K,h_Q}}
 \ge 
\frac{16C_Q\overline B_Y^2}
{c_\ell^2(1-\gamma)^2}\epsilon^{-4}.
\end{align*}
Since
\(
\norm{\widehat Q_0}_\infty\le B_Q,
\quad
\norm{Q_\chi^*}_\infty\le B_Q,
\)
we have
\(
\norm{\widehat Q_0-Q_\chi^*}_\infty\le2B_Q.
\)
Thus it is sufficient to choose
\[
T=
\left\lceil
\frac{
\log\!\big((8B_Q/\epsilon)\vee1\big)
}{
\log(1/\gamma)
}
\right\rceil .
\]
With this choice, Algorithm~\ref{alg:main} satisfies
\(
\E\norm{\widehat Q_T-Q_\chi^*}_\infty\le\epsilon.
\)

Consequently, every prescribed \(\epsilon\in(0,1)\) satisfying
\eqref{eq:approximation-floor-condition} can be achieved with
\[
T(N+K)
\le
C_{\rm comp}\,
\epsilon^{-6}\log^3\frac{e}{\epsilon},
\]
where \(C_{\rm comp}\) depends only on the fixed problem, mixing,
function-class, and design parameters. Hence the sufficient budget is
\(\widetilde O(\epsilon^{-6})\) over the admissible target accuracies.

\end{corollary}

\begin{proof}
For \(0<\ell\le1\),
\[
G_\ell\le\frac{\overline G}{\ell^2},
\qquad
B_\varphi(\ell)\le\frac{\overline B_\varphi}{\ell},
\qquad
B_Y(\ell)\le\frac{\overline B_Y}{\ell}.
\]
The two requirements on \(N\) make the two terms in
\(\gamma\overline\varepsilon_{\rm var}/
[\mu_\phi(1-\gamma)]\)
at most \(\epsilon/8\) each. The requirement on \(K\) gives
\(\varepsilon_Q/(1-\gamma)\le\epsilon/4\).
Condition~\eqref{eq:approximation-floor-condition} controls the
Bellman-, dual-, and scale-function approximation terms together with
the explicit floor bias. Finally, since
\(\norm{\widehat Q_0-Q_\chi^*}_\infty\le2B_Q\), the stated choice of
\(T\) makes the initial-error term at most \(\epsilon/4\).
\end{proof}

\begin{remark}
    Under exact realizability,
\(\mathcal E_Q=\mathcal E_\eta=\mathcal E_{\rm sc}=0\),
condition~\eqref{eq:approximation-floor-condition} holds for every
prescribed \(\epsilon\in(0,1)\) whenever
\(c_\ell\le\mu_\phi(1-\gamma)/(2\gamma c_\delta)\).
Here \(\ell=c_\ell\epsilon\) is fixed within each run; the complexity
scaling compares separately tuned runs. A run with fixed
\(\ell>0\) retains the \(O(\ell)\) floor term in
Theorem~\ref{thm:main}.
\end{remark}

\begin{remark}[Complexity under a relaxed gain]
\label{rem:slow-q}
At the boundary \(p_Q=1\), the extra logarithm in
Remark~\ref{rem:q-general-gain} leaves the same
\(\widetilde O(\epsilon^{-6})\) total sample-complexity order.  If
\(0<p_Q<1\), the Q-stage requirement becomes
\[
K=\widetilde O(\epsilon^{-4/p_Q})
\quad\text{when}\quad\ell=\Theta(\epsilon).
\]
Thus Stage~1 continues to determine the
\(\widetilde O(\epsilon^{-6})\) exponent whenever
\(p_Q\ge2/3\).  For \(p_Q<2/3\), the total exponent becomes
\(\max\{6,4/p_Q\}\).  This is the precise cost of removing the
clean-rate condition \(2\mu_\phi\beta_0>1\).
\end{remark}

Next, we show how a greedy policy with respect to the returned $\widehat Q_T$ performs in the robust setting. For a deterministic policy \(\pi\), let \(Q_\chi^\pi\) be the fixed
point of
\((\mathcal T_\chi^\pi Q)(i)=r(i)+\gamma\sigma_i^\chi
(Q(\cdot,\pi(\cdot)))\), and set
\(V_\chi^\pi(s):=Q_\chi^\pi(s,\pi(s))\) and
\(V_\chi^*:=V_{Q_\chi^*}\). $V_\chi^\pi$ and $V_\chi^*$ are the robust value function evaluated at policy $\pi$ and the optimal robust value function, respectively.

\begin{corollary}[Greedy robust policy]
\label{cor:greedy}
Let
\(\widehat\pi(s)\in\argmax_{a\in\cA}\widehat Q_T(s,a)\), with ties
resolved by a fixed deterministic rule. Then, almost surely,
\begin{align}
\norm{V_\chi^{\widehat\pi}-V_\chi^*}_\infty
&\le \frac{2}{1-\gamma}
\norm{\widehat Q_T-Q_\chi^*}_\infty.
\label{eq:policy-bound}
\end{align}
\end{corollary}

\begin{proof}
Set \(e=\norm{\widehat Q_T-Q_\chi^*}_\infty\),
\(q=\norm{Q_\chi^{\widehat\pi}-Q_\chi^*}_\infty\), and
\(v=\norm{V_\chi^{\widehat\pi}-V_\chi^*}_\infty\).
Greediness gives
\(0\le V_\chi^*(s)-Q_\chi^*(s,\widehat\pi(s))\le2e\).
The Bellman equations and the one-Lipschitz property of
\(\sigma_i^\chi\) give \(q\le\gamma v\), while adding and subtracting
\(Q_\chi^*(s,\widehat\pi(s))\) gives \(v\le2e+q\).
Solving these inequalities proves \eqref{eq:policy-bound}.
\end{proof}

\subsection{Interpreting the Approximation Requirement}
\label{sec:approximation}
For fixed function classes and \(\ell\), increasing \(N\) and \(K\)
reduces the statistical terms, while increasing \(T\) reduces the outer
transient. The errors
\(\mathcal E_Q,\mathcal E_\eta,\mathcal E_{\rm sc}\) and the floor term
determine the residual neighborhood certified by
Theorem~\ref{thm:main}. The Bellman
error $\mathcal E_Q$ measures approximate closure of the Q-function
class under the clipped robust Bellman update; representability of
$Q_\chi^*$ alone does not imply $\mathcal E_Q=0$.

The dual error measures the loss from representing the pointwise scalar
optimizers by one shared parameter. If \(\eta_V^*:=(\eta_V^*(i))_i\) collects pointwise scalar-dual
maximizers, the \((1+c_\delta)\)-Lipschitz property of
\(F_V^\chi(i;\cdot)\) gives 
\begin{equation}
\mathcal A_\eta(V,\nu)
\leq
(1+c_\delta)\Dnorm{\Psi\nu-\eta_V^*}.
\label{eq:dual-ordinary-approx}
\end{equation}
Thus, $\mathcal E_\eta$ is controlled by an ordinary weighted
function-approximation error. Similarly, $\mathcal E_{\rm sc}$
measures approximation of the unshifted scale $s^{V,\nu}$.
The term $\mathcal E_{\rm sc}^2/\ell$ makes the floor tradeoff
explicit: decreasing $\ell$ reduces the floor bias but enlarges the contribution of scale misspecification and the
statistical bounds.

The tabular one-hot specialization gives an exact-realizability
benchmark. Taking
$\phi(i)=\psi(i)=\chi(i)=e_i$,
$M=|\mathcal S||\mathcal A|$, and
$d_{\min}=\min_i d_i$, all three approximation errors vanish whenever
\[
R_\nu\geq
\sqrt M\,B_Q(1+\delta^{-1/2}),
\qquad
R_\Theta\geq
\sqrt M\,B_Q(2+\delta^{-1/2}),
\]
and one may take any \(0<\mu_\phi\le d_{\min}\) with \(\mu_\phi<1\). The tabular experiment therefore
isolates the finite-sample and floor effects, whereas the neural
experiment does not assume exact realizability.

\section{Finite-Time Analysis}
\label{sec:proofs}

We first collect the trajectory notation and the two consequences of
Assumption~\ref{ass:mixing} used repeatedly below. Let \(\mathcal F_k\) be the \(\sigma\)-field generated by the trajectory
through \(I_k\) and all algorithmic iterates formed before
\(S_{k+1}\) is drawn. The Markov property gives, for every $k\geq L\geq0$,
\begin{equation}
\operatorname{Law}(I_k\mid\mathcal F_{k-L})
=
P_b^L(I_{k-L},\cdot),
\label{eq:relative-markov}
\end{equation}
and hence
\begin{equation}
\left\|
\operatorname{Law}(I_k\mid\mathcal F_{k-L})-d
\right\|_{\rm TV}
\leq
C_{\rm mix}\rho_{\rm mix}^L.
\label{eq:conditional-mixing}
\end{equation}
More generally, let
$h:\Omega\times(\mathcal S\times\mathcal A)\to\mathbb R$ be a random
map such that $h(\cdot,i)$ is $\mathcal F_{k-L}$-measurable for every
$i$, and suppose $\sup_i|h(i)|\leq B_h$ almost surely. Then
\begin{equation}
\left|
\E[h(I_k)\mid\mathcal F_{k-L}]
-\sum_i d_i h(i)
\right|
\leq
2B_hC_{\rm mix}\rho_{\rm mix}^L.
\label{eq:function-mixing}
\end{equation}

For outer block $t$, define the global indices
\begin{equation}
b_t=t(N+K),
\qquad
q_t=b_t+N.
\label{eq:block-indices}
\end{equation}
Stage~1 uses indices $b_t,\ldots,q_t-1$, and Stage~2 uses indices
$q_t,\ldots,b_{t+1}-1$. Let
\begin{equation}
\mathcal B_t:=\mathcal F_{b_t},
\qquad
\mathcal H_{t,m}:=\mathcal F_{q_t+m}.
\label{eq:block-filtrations}
\end{equation}
The quantities
$V_t,\bar\nu_t,\bar\Theta_t,\theta_{t,m}$, and $I_{q_t+m}$ are
$\mathcal H_{t,m}$-measurable, whereas
\[
S_{q_t+m+1}\mid\mathcal H_{t,m}
\sim P_0(\cdot\mid I_{q_t+m}).
\]
Consequently,
\begin{equation}
\E[Y_{t,m}\mid\mathcal H_{t,m}]
=
\overline Y_{V_t,\bar\nu_t,\bar\Theta_t}(I_{q_t+m}).
\label{eq:q-conditional-mean}
\end{equation}

The proof mirrors the three levels of Algorithm~\ref{alg:main}.
Theorem~\ref{thm:stage1} analyzes Stage~1 as projected stochastic
ascent on the frozen variational objective. After the learned dual and
scale parameters are frozen, Theorem~\ref{thm:q-stage} analyzes
Stage~2 as a linear stochastic-approximation recursion toward the
projected variational target. Lemma~\ref{lem:target-discrepancy}
converts the Stage~1 objective gap into a robust-target discrepancy,
and Proposition~\ref{prop:block-error} combines this discrepancy with
the Q-stage error. Finally, Lemma~\ref{lem:outer-contraction}
propagates the one-block error across blocks using contraction of the
exact robust Bellman operator.

\subsection{Stage 1: variational objective gap}

Although the one-transition gradient is conditionally unbiased for the
currently visited state--action pair, the iterate at time $k$ depends
on the recent history of the behavior chain. We therefore compare it
with an iterate sufficiently far in the past to use geometric mixing.
For a Stage~1 block of length $N$, define
\begin{equation}
L_N
:=
\left\lceil
\frac{\log\!\big(C_{\rm mix}(N+1)\vee1\big)}
     {\log(1/\rho_{\rm mix})}
\right\rceil .
\label{eq:LN}
\end{equation}
By construction,
\[
C_{\rm mix}\rho_{\rm mix}^{L_N}\leq\frac{1}{N+1}.
\]
The proof inserts the lagged iterate $w_{k-L_N}$: mixing controls its
sample-to-stationary discrepancy, while bounded gradients control the
difference between $w_k$ and $w_{k-L_N}$.

\begin{theorem}[Stage~1 objective gap]
\label{thm:stage1}
For every outer block \(t\), every integer \(N\ge1\), and every
\(\alpha_N>0\), almost surely,
\begin{align}
&\E\!\left[
J_{V_t,\ell}^{\rm PSD,*}
-J_{V_t,\ell}(\bar\nu_t,\bar\Theta_t)
\;\middle|\;\mathcal B_t
\right]
\le\varepsilon_{\rm var}(N,\ell),
\label{eq:stage1-explicit}\\
\varepsilon_{\rm var}(N,\ell)
&:=
\frac{R_{\mathcal W}^2}{2\alpha_NN}
+\alpha_NG_\ell^2\left(\frac12+2L_N\right)
\nonumber\\
&
+4B_\varphi(\ell)C_{\rm mix}\rho_{\rm mix}^{L_N}
+\frac{4B_\varphi(\ell)(L_N\wedge N)}{N}.
\label{eq:eps-var-exact}
\end{align}
For the choice of \(\alpha_N\) in \eqref{eq:main-stepsizes},
\begin{equation}
\varepsilon_{\rm var}(N,\ell)
\le
\overline\varepsilon_{\rm var}(N,\ell).
\label{eq:eps-var-simplified}
\end{equation}
\end{theorem}

\begin{proof}
Fix block \(t\), condition on \(\mathcal B_t\), and suppress \(t\) and
the frozen target \(V_t\) from the notation.  Write
\[
w_k=(\nu_k,\Theta_k),
\qquad
J(w)=J_{V_t,\ell}(w).
\]

Since \(V_t\) is \(\mathcal B_t\)-measurable,
\(\mathcal W\) is compact, and
\((V,w)\mapsto J_{V,\ell}(w)\) is jointly measurable in \(V\) and
continuous in \(w\), the measurable maximum theorem yields a
\(\mathcal B_t\)-measurable selection
\begin{equation}
w^*
\in
\argmax_{w\in\mathcal W}J_{V_t,\ell}(w).
\label{eq:measurable-comparator}
\end{equation}

The Stage~1 initialization and parameter constraints give
\(\norm{w_0-w^*}\leq R_{\mathcal W}\).

Let \(Z_k=(I_{b_t+k},S_{b_t+k+1})\), and let \(g_k\) denote the
stacked sample gradient in \eqref{eq:sample-gradients} evaluated at
\((w_k,Z_k)\). Projection nonexpansiveness gives

\begin{align*}
\norm{w_{k+1}-w^*}^2
&\le
\norm{w_k-w^*}^2
+\alpha_N^2\norm{g_k}^2
\nonumber\\
&\quad
-2\alpha_N\inner{g_k}{w^*-w_k}.
\end{align*}
The sample objective is concave, so
\[
\widehat\varphi(w^*;Z_k)-\widehat\varphi(w_k;Z_k)
\le
\inner{g_k}{w^*-w_k}.
\]
Using \(\norm{g_k}\le G_\ell\), summing from \(0\) to \(N-1\), and
telescoping,
\begin{equation}
\sum_{k=0}^{N-1}
\bigl[
\widehat\varphi(w^*;Z_k)-\widehat\varphi(w_k;Z_k)
\bigr]
\le
\frac{R_{\mathcal W}^2}{2\alpha_N}
+\frac{\alpha_NG_\ell^2N}{2}.
\label{eq:pathwise-regret}
\end{equation}

It remains to compare the sampled objectives with the stationary
population objective $J(w_k)$ defined in
\eqref{eq:population-objective}. Define
\[
\bar\varphi(w;i)
:=
\E[\widehat\varphi(w;i,S')\mid i].
\]

On \(\mathcal W\), \(w\mapsto\bar\varphi(w;i)\) is
\(G_\ell\)-Lipschitz and
\(|\widehat\varphi(w;i,s')|,|\bar\varphi(w;i)|
\le B_\varphi(\ell)\).

For \(k\geq L_N\), projection nonexpansiveness and
\(\norm{g_j}\leq G_\ell\) give
\begin{equation}
\norm{w_k-w_{k-L_N}}
\leq\alpha_NG_\ell L_N.
\label{eq:stage1-lag-distance}
\end{equation}

Set \(\iota_k:=I_{b_t+k}\).  Insert and subtract the lagged
iterate $w_{k-L_N}$:
\begin{align*}
&\bar\varphi(w_k;\iota_k)-J(w_k)
=
\left[\bar\varphi(w_k;\iota_k)
-\bar\varphi(w_{k-L_N};\iota_k) \right]
\\
&+
\left[\bar\varphi(w_{k-L_N};\iota_k)
-J(w_{k-L_N})\right] + \left[
J(w_{k-L_N})-J(w_k) \right].
\end{align*}
The first and third terms have absolute value at most
\(G_\ell\norm{w_k-w_{k-L_N}}\).  Conditional on the lagged
filtration, \(w_{k-L_N}\) is fixed.  Equations
\eqref{eq:function-mixing} and \eqref{eq:stage1-lag-distance} therefore
give
\begin{align}
&\left|
\E\!\left[
\widehat\varphi(w_k;Z_k)-J(w_k)
\;\middle|\;\mathcal B_t
\right]
\right|
\nonumber\\
&\qquad\le
2\alpha_NG_\ell^2L_N
+2B_\varphi(\ell)C_{\rm mix}\rho_{\rm mix}^{L_N}.
\label{eq:iterate-bias}
\end{align}
Here we also used
\[
\E[\widehat\varphi(w_k;Z_k)\mid\mathcal F_{b_t+k}]
=\bar\varphi(w_k;I_{b_t+k}).
\]
For the \(\mathcal B_t\)-measurable comparator \(w^*\), no lagged
iterate is needed; for \(k\ge L_N\), the same mixing argument gives
\begin{equation}
\left|
\E\!\left[
\widehat\varphi(w^*;Z_k)-J(w^*)
\;\middle|\;\mathcal B_t
\right]
\right|
\le
2B_\varphi(\ell)C_{\rm mix}\rho_{\rm mix}^{L_N}.
\label{eq:comparator-bias}
\end{equation}

For \(0\le k<L_N\wedge N\), each of the two sample-to-population
discrepancies is at most \(2B_\varphi(\ell)\); their combined
contribution is therefore at most \(4B_\varphi(\ell)\) per iteration.
Taking conditional expectation in \eqref{eq:pathwise-regret}, using
\eqref{eq:iterate-bias}--\eqref{eq:comparator-bias} on the remaining
indices, and dividing by \(N\), gives
\[
\frac1N\sum_{k=0}^{N-1}
\E[J(w^*)-J(w_k)\mid\mathcal B_t]
\le\varepsilon_{\rm var}(N,\ell).
\]
Since \(J\) is concave and
\(\bar w=N^{-1}\sum_{k=0}^{N-1}w_k\),
\[
J(\bar w)\ge\frac1N\sum_{k=0}^{N-1}J(w_k),
\]
which proves \eqref{eq:stage1-explicit}.

Finally,
\begin{align*}
C_{\rm mix}\rho_{\rm mix}^{L_N}
&\le\frac1{N+1},
\nonumber\\
L_N
&\le\kappa_{\rm mix}(1+\log(N+1))
\le\kappa_{\rm mix}\Lambda_N.
\end{align*}
Substitution of
\(\alpha_N=R_{\mathcal W}/(G_\ell\sqrt{N\Lambda_N})\)
into \eqref{eq:eps-var-exact} yields
\eqref{eq:eps-var-simplified}.
\end{proof}

\subsection{Stage 2: Q-stage linear stochastic approximation}
\label{subsec:q-stage}

Once $(V,\nu,\Theta)$ is frozen, Stage~2 regresses the corresponding
one-transition targets onto the Q-feature space. Its population target
is the $D$-weighted least-squares parameter
\begin{equation}
\theta^{\rm var}_{V,\nu,\Theta}
:=
(\Phi^\top D\Phi)^{-1}\Phi^\top D
\overline Y_{V,\nu,\Theta}.
\label{eq:theta-var}
\end{equation}
For outer block $t$, abbreviate
\[
\theta_t^{\rm var}
:=
\theta^{\rm var}_{V_t,\bar\nu_t,\bar\Theta_t},
\qquad
\E_t[\cdot]:=\E[\cdot\mid\mathcal H_{t,0}].
\]
Thus, $\theta_t^{\rm var}$ is the fixed point of the mean linear
recursion within block $t$. It is not yet the ideal robust Bellman
parameter because $(\bar\nu_t,\bar\Theta_t)$ has been estimated from
finite data; Proposition~\ref{prop:block-error} handles that
discrepancy separately.

\begin{theorem}[Q-stage error]
\label{thm:q-stage}
Suppose Assumptions~\ref{ass:features} and~\ref{ass:mixing} hold.
Fix any $\mathcal H_{t,0}$-measurable
$(V_t,\bar\nu_t,\bar\Theta_t)$ satisfying
$\norm{V_t}_\infty\le B_Q$ and
$(\bar\nu_t,\bar\Theta_t)\in\mathcal W$.
Initialize $\theta_{t,0}=0$ and use
\begin{equation}
\beta_m=\frac{\beta_0}{m+h_Q},
\qquad
h_Q\ge\max\{1,2\beta_0\},
\qquad
p_Q>1.
\label{eq:q-stage-stepsize}
\end{equation}
Then, for every $K\ge1$, almost surely,
\begin{align}
\E_t\norm{\theta_{t,K}-\theta_t^{\rm var}}_2^2
&\le
B_Y(\ell)^2C_Q\mathfrak R_Q(K),
\label{eq:q-mse}\\
\E_t\norm{\theta_{t,K}-\theta_t^{\rm var}}_2
&\le
B_Y(\ell)\sqrt{C_Q\mathfrak R_Q(K)}.
\label{eq:q-l1}
\end{align}
The constant $C_Q$ given in Equation \eqref{eq:CQ-explicit} is independent of $t,K,\ell$, and the frozen
inputs.
\end{theorem}

We next collect the constants needed to make $C_Q$ explicit. They
encode the feature conditioning, gain, offset, and mixing behavior; the
magnitude of the frozen target has already been extracted through
$B_Y(\ell)$.

 Put
\begin{align}
q_{\rm mix}
&:=\log(1/\rho_{\rm mix}),
&
C_{\rm mix}^Q
&:=
\max\left\{
1,\frac{C_{\rm mix}}{\rho_{\rm mix}}
\right\},
\nonumber\\
\overline C_L
&:=
\left(C_{\rm mix}^Q/\beta_0\right)\vee1,
&
c_L
&:=
1+\frac{1+\log\overline C_L}{q_{\rm mix}},
\nonumber\\
\kappa_Q&:=2+8\beta_0.
\label{eq:q-lag-constants}
\end{align}
and
\begin{equation}
L_m^Q
:=
\max\left\{
1,
\left\lceil
\frac{
\log\!\left(
C_{\rm mix}^Q(m+h_Q)/\beta_0\vee1
\right)
}{
q_{\rm mix}
}
\right\rceil
\right\}.
\label{eq:q-adaptive-lag}
\end{equation}
Then
\[
C_{\rm mix}^Q\rho_{\rm mix}^{L_m^Q}
\le\beta_m,
\qquad
L_m^Q
\le
c_L[1+\log(m+h_Q)].
\]
Define
\begin{align}
m_Q&:=
\left\lceil
\max\{h_Q,4(\kappa_Qc_L)^2\}
\right\rceil,
&
n_{m_Q}&:=m_Q+h_Q,
\nonumber\\
c_{\rm SA}&:=228,
&
D_{Q,1}&:=8c_{\rm SA}\beta_0^2,
\nonumber\\
D_{Q,0}&:=2c_{\rm SA}\beta_0^2
\left(2/\mu_\phi+1\right)^2,
\label{eq:q-drift-constants}
\end{align}
\begin{align}
U_Q&:=
\left[
\frac1{\mu_\phi}
+\beta_0\left(1+\log\frac{n_{m_Q}}{h_Q}\right)
\right]^2,
\nonumber\\
C_\Pi&:=
\exp\left\{
D_{Q,1}c_L\frac{3+2\log n_{m_Q}}{n_{m_Q}}
\right\}.
\label{eq:q-product-constant}
\end{align}
Here \(c_{\rm SA}\) is the universal constant from the Markovian-SA
drift lemma used below; it is not a problem parameter.  For
$p_Q:=2\mu_\phi\beta_0>1$, set
\begin{equation}
C_Q
:=
C_\Pi\left[
U_Qn_{m_Q}^{p_Q}
+2^{p_Q}c_LD_{Q,0}
\left(1+\frac1{p_Q-1}\right)
\right].
\label{eq:CQ-explicit}
\end{equation}

\begin{proof}
Condition on \(\mathcal H_{t,0}\). Under this conditional law the
frozen inputs are deterministic and the future behavior chain retains
its original kernel. Write \(B=B_Y(\ell)\) and set
$x_m=\theta_{t,m}/B$ and
$x_t^{\rm var}=\theta_t^{\rm var}/B$.  With
$\Xi_m=(I_{q_t+m},S_{q_t+m+1})$, define
\begin{equation}
F_t(x;(i,s'))
:=
\bigl(I-\phi(i)\phi(i)^\top\bigr)x
+\frac{Y_t(i,s')}{B}\phi(i),
\label{eq:q-SA-map}
\end{equation}
where
\[
Y_t(i,s')
:=
r(i)+\gamma
\widehat\varphi_{V_t,\ell}
(\bar\nu_t,\bar\Theta_t;i,s').
\]
The Q-stage update can then be written as
\begin{equation}
x_{m+1}
=x_m+\beta_m\bigl(F_t(x_m;\Xi_m)-x_m\bigr).
\label{eq:q-SA-form}
\end{equation}

Under this conditional law, \((\Xi_m)\) is Markov with kernel
\(P_\Xi\); the following bounds are uniform in its initial
distribution. Its stationary distribution is
\[
\widetilde d(i,s')
=
d_iP_0(s'\mid i).
\]
For \(L\ge1\), conditioning on
\(\Xi_0=(i_0,s_1)\) leaves \(L-1\) transitions of the
state--action chain after the action at \(s_1\) is sampled. Hence,
by convexity of total variation and contraction under the kernel
\(i\mapsto(i,S')\),
\[
\sup_{\xi}
\left\|
P_\Xi^L(\xi,\cdot)-\widetilde d
\right\|_{\rm TV}
\le
C_{\rm mix}\rho_{\rm mix}^{L-1}
\le
C_{\rm mix}^Q\rho_{\rm mix}^{L}.
\]
The case \(L=0\) is covered by \(C_{\rm mix}^Q\ge1\).
Therefore, the augmented process satisfies the geometric mixing
condition required by the Markovian-SA drift lemma with constants
\((C_{\rm mix}^Q,\rho_{\rm mix})\).

Let
\(A_\phi:=\Phi^\top D\Phi\) and
\(\overline Y_t:=
\overline Y_{V_t,\bar\nu_t,\bar\Theta_t}\).
The stationary mean map of \(F_t\) is
\[
\overline F_t(x)
=(I-A_\phi)x+\frac{\Phi^\top D\overline Y_t}{B},
\]
whose unique fixed point is $x_t^{\rm var}$.  Since
$\mu_\phi I\preceq A_\phi\preceq I$ and $|Y_t(i,s')|\le B$,
\begin{align}
\norm{\overline F_t(x)-\overline F_t(z)}_2
&\le(1-\mu_\phi)\norm{x-z}_2,
\nonumber\\
\norm{F_t(x;\xi)-F_t(z;\xi)}_2
&\le\norm{x-z}_2,
\nonumber\\
\norm{F_t(0;\xi)}_2&\le1,
\qquad
\norm{x_t^{\rm var}}_2\le\mu_\phi^{-1}.
\label{eq:q-SA-verification}
\end{align}

Apply the pre-absorption one-step inequality in the proof of
\cite[Prop.~2]{chen2024lyapunov} with
\(\norm{\cdot}_c=\norm{\cdot}_s=\norm{\cdot}_2\), Euclidean
smoothness constant one, and the proposition's separate
additive-noise term set to zero. In the source notation,
\[
M(z)=\frac14\norm{z}_2^2,\quad
\ell_{cm}=u_{cm}=\sqrt2,\quad
A_{\rm SA}=2, B_{\rm SA}=1,
\]
while the contraction factor is \(1-\mu_\phi\), so
\(\phi_2=\mu_\phi\) and \(\phi_3=228=:c_{\rm SA}\).

Thus, if a lag \(L\leq m\) satisfies
\[
C_{\rm mix}^Q\rho_{\rm mix}^{L}\leq\beta_m,
\qquad
\Delta_m(L):=\sum_{j=m-L}^{m-1}\beta_j\leq\frac18,
\]
then, since \(M(z)=\norm{z}_2^2/4\), multiplying the resulting
inequality by four and defining
\[
a_m:=\E_t\norm{x_m-x_t^{\rm var}}_2^2
\]
gives
\begin{align}
a_{m+1}
&\leq
\left(
1-2\mu_\phi\beta_m
+4c_{\rm SA}\beta_m\Delta_m(L)
\right)a_m
\nonumber\\
&\quad+
c_{\rm SA}(2/\mu_\phi+1)^2
\beta_m\Delta_m(L).
\label{eq:q-unabsorbed-drift}
\end{align}
We retain the lag perturbation explicitly and control it in the
product unrolling below.

We now verify the two lag conditions for $L=L_m^Q$. For
$m\geq m_Q$, put $n_m=m+h_Q$. Since
$m_Q\geq h_Q$ and $m_Q\geq4(\kappa_Qc_L)^2\geq16$, we have
$n_m\leq2m$ and $1+\log n_m\leq\sqrt{n_m}$. Therefore,
\[
L_m^Q
\leq c_L\sqrt{n_m}
\leq\frac{n_m}{2}
\leq m.
\]
Moreover, for $m-L_m^Q\leq j\leq m-1$,
$j+h_Q\geq n_m-L_m^Q\geq n_m/2$, and hence
\begin{equation}
\Delta_m
:=
\sum_{j=m-L_m^Q}^{m-1}\beta_j
\leq
\frac{2\beta_0L_m^Q}{n_m}
\leq
\frac{2\beta_0c_L}{\sqrt{n_m}}
\leq
\frac{\beta_0}{\kappa_Q}
\leq
\frac18.
\label{eq:q-lag-mass}
\end{equation}
The penultimate inequality uses
$\sqrt{n_m}\geq2\kappa_Qc_L$, and the last follows from
$\kappa_Q=2+8\beta_0$. Together with
$C_{\rm mix}^Q\rho_{\rm mix}^{L_m^Q}\leq\beta_m$, these bounds permit
the use of \eqref{eq:q-unabsorbed-drift}. Substituting
$\beta_m=\beta_0/n_m$ and
$\Delta_m\leq2\beta_0L_m^Q/n_m$ gives
\begin{align}
a_{m+1}
&\leq
\left[
1-\frac{p_Q}{m+h_Q}
+D_{Q,1}\frac{L_m^Q}{(m+h_Q)^2}
\right]a_m
\nonumber\\
&\quad+
D_{Q,0}\frac{L_m^Q}{(m+h_Q)^2}.
\label{eq:q-scalar-explicit}
\end{align}

Write
\[
\phi_m:=\phi(I_{q_t+m}),
\qquad
y_m:=\frac{Y_t(I_{q_t+m},S_{q_t+m+1})}{B}.
\]
Then \eqref{eq:q-SA-form} becomes
\[
x_{m+1}
=(I-\beta_m\phi_m\phi_m^\top)x_m
+\beta_m y_m\phi_m.
\]
Since
\(\beta_m\norm{\phi_m}_2^2\le\beta_0/h_Q\le1/2\),
\[
\norm{I-\beta_m\phi_m\phi_m^\top}_{\rm op}\le1,
\]
while \(|y_m|\le1\) and \(\norm{\phi_m}_2\le1\).  Therefore,
\begin{align*}
    \norm{x_m-x_t^{\rm var}}_2
& \le \norm{x_t^{\rm var}}_2+\sum_{j=0}^{m-1}\beta_j\\
& \le
\frac1{\mu_\phi}
+\beta_0\left[
1+\log\left(\frac{m+h_Q}{h_Q}\right)
\right].
\end{align*}
Since \(m+h_Q\le n_{m_Q}\) for \(m\le m_Q\), it follows that
\begin{equation}
a_m\le U_Q,
\qquad 0\le m\le m_Q.
\label{eq:q-prefix-bound}
\end{equation}

For \(K>m_Q\), set \(n_K:=K+h_Q\) and, for \(m\ge m_Q\),
\[
c_m:=1-\frac{p_Q}{n_m}
+D_{Q,1}\frac{L_m^Q}{n_m^2}.
\]
Since \(n_m\ge2h_Q\ge4\beta_0\ge2p_Q\), we have \(c_m\ge1/2\), so
\eqref{eq:q-scalar-explicit} may be unrolled. Integral comparison,
\(1+x\le e^x\), and
\(\sum_{j=r}^{K-1}n_j^{-1}\ge\log(n_K/n_r)\) give
\begin{align*}
\sum_{j=m_Q}^\infty\frac{L_j^Q}{n_j^2}
&\le c_L\frac{3+2\log n_{m_Q}}{n_{m_Q}},\\
\prod_{j=r}^{K-1}c_j
&\le C_\Pi\left(\frac{n_r}{n_K}\right)^{p_Q},\\
\sum_{r=m_Q}^{K-1}n_r^{p_Q-2}
&\le\left(1+\frac1{p_Q-1}\right)n_K^{p_Q-1}.
\end{align*}
Hence, using \(L_r^Q\le c_L(1+\log n_K)\) and
\(n_{r+1}\le2n_r\),
\begin{align*}
a_K
&\le \frac{C_\Pi U_Qn_{m_Q}^{p_Q}}{n_K^{p_Q}}
 +\frac{C_\Pi D_{Q,0}}{n_K^{p_Q}}
 \sum_{r=m_Q}^{K-1}\frac{L_r^Qn_{r+1}^{p_Q}}{n_r^2}\\
&\le \frac{C_\Pi U_Qn_{m_Q}^{p_Q}}{n_K^{p_Q}}\\
& +C_\Pi2^{p_Q}c_LD_{Q,0}
\left(1+\frac1{p_Q-1}\right)
\frac{1+\log n_K}{n_K}.
\end{align*}
Together with \eqref{eq:q-prefix-bound} and the definitions of \(C_Q\)
and \(\mathfrak R_Q\), this yields
\(a_K\le C_Q\mathfrak R_Q(K)\) for every \(K\ge1\). Multiplication by $B^2$ proves \eqref{eq:q-mse}; Jensen's inequality
gives \eqref{eq:q-l1}.
\end{proof}

\begin{remark}[Q-stage rates for an arbitrary gain]
\label{rem:q-general-gain}
The derivation through \eqref{eq:q-scalar-explicit} only requires
$\beta_0>0$ and $h_Q\ge\max\{1,2\beta_0\}$; $p_Q>1$ is used only in
the power-sum bound.  The same unrolling gives normalized mean-square
rates $O((K+h_Q)^{-p_Q})$ for $0<p_Q<1$ and
$O(\log^2(K+h_Q)/(K+h_Q))$ for $p_Q=1$, with constants independent of
$K,N,T,\ell,\gamma$, and $\delta$.
\end{remark}

\begin{remark}[Dependence on the main problem parameters]
\label{rem:parameter-dependence}
For $0<\ell\leq1$, fix the approximation radii and mixing parameters.
One admissible tuning is
\[
\beta_0=\mu_\phi^{-1},
p_Q=2,
h_Q=
\left\lceil
4(\kappa_Qc_L)^2
\left[1+\log\!\bigl(4(\kappa_Qc_L)^2\bigr)\right]
\right\rceil .
\]
Then $m_Q=h_Q$, $n_{m_Q}=2h_Q$, and the constants defined above satisfy
\[
h_Q=\widetilde O(\mu_\phi^{-2}),
\qquad
C_\Pi=O(1),
\qquad
\sqrt{C_Q}=\widetilde O(\mu_\phi^{-3}).
\]
Consequently, after suppressing logarithmic factors and constants
depending only on the fixed radii and mixing parameters, the leading
Q-stage statistical term in Theorem~\ref{thm:main} scales as
\[
\widetilde O\!\left(
\frac{\mu_\phi^{-3}}{\sqrt{K+h_Q}}
\left[
\frac{1}{1-\gamma}
+
\frac{\gamma \sqrt{1+\delta}}{\ell(1-\gamma)^3}
\right]
\right),
\]
while the leading Stage~1 term scales as
\[
\widetilde O\!\left(
\frac{\gamma \sqrt{1+\delta}}{\mu_\phi(1-\gamma)^3}
\left[
\frac{1}{\ell^2\sqrt N}
+\frac{1}{\ell N}
\right]
\right).
\]
The remaining approximation and floor terms are displayed explicitly
in \eqref{eq:main-bound}. This tuning is sufficient rather than
optimal; if the approximation radii depend on $(\gamma,\delta)$, their
additional dependence enters through $H$, $R_{\mathcal W}$, and the
approximation errors.
\end{remark}

\subsection{From variational objective gap to one-block error}

The Stage~1 analysis controls an averaged variational objective gap.
We next show that this objective gap controls the weighted discrepancy
between the variational target and the exact robust target, and hence
the parameter error incurred within one outer block.

\begin{lemma}[One-sided target discrepancy]
\label{lem:target-discrepancy}
For every bounded \(V\), every feasible \((\nu,\Theta)\), and every
state--action pair \(i\),
\begin{equation}
f_{V,\ell}(i;\nu,\Theta)
\le
\sigma_i^\chi(V).
\label{eq:one-sided}
\end{equation}
Moreover, if \(V=V_Q\) for some \(Q\in\cQ\), then
\begin{align}
&\sum_i d_i
\left|
\sigma_i^\chi(V)
-f_{V,\ell}(i;\nu,\Theta)
\right|
\nonumber\\
&\quad\le
J_{V,\ell}^{\mathrm{PSD},*}
-J_{V,\ell}(\nu,\Theta)
+\mathcal E_\eta
+\frac{c_\delta}{2}
\left(
\mathcal E_{\rm sc}
+\frac{\mathcal E_{\rm sc}^2}{\ell}
\right)
+\frac{c_\delta\ell}{2}.
\label{eq:weighted-discrepancy}
\end{align}
\end{lemma}

\begin{proof}
For every scalar \(\eta\) and every \(u>0\),
\eqref{eq:qol} gives
\[
\eta-\frac{c_\delta}{2}
\left(
\frac{Z_V(i;\eta)}{u}+u
\right)
\le
\eta-c_\delta\sqrt{Z_V(i;\eta)}
\le
\sigma_i^\chi(V).
\]
Substituting
\[
\eta=\psi(i)^\top\nu,
\qquad
u=\ell+g_\Theta(i)
\]
proves \eqref{eq:one-sided}. Consequently, the difference inside
the absolute value in \eqref{eq:weighted-discrepancy} is nonnegative,
and
\begin{align*}
&\sum_i d_i
\left|
\sigma_i^\chi(V)
-f_{V,\ell}(i;\nu,\Theta)
\right|
=
\sum_i d_i\sigma_i^\chi(V)
-J_{V,\ell}(\nu,\Theta)\\
&\quad=
\left[
\sum_i d_i\sigma_i^\chi(V)
-J_{V,\ell}^{\mathrm{PSD},*}
\right]
+
\left[
J_{V,\ell}^{\mathrm{PSD},*}
-J_{V,\ell}(\nu,\Theta)
\right].
\end{align*}
Since \(V=V_Q\) for some \(Q\in\cQ\),
Proposition~\ref{prop:fa-floor-decomposition} bounds the first
bracket by
\[
\mathcal E_\eta
+\frac{c_\delta}{2}
\left(
\mathcal E_{\rm sc}
+\frac{\mathcal E_{\rm sc}^2}{\ell}
\right)
+\frac{c_\delta\ell}{2}.
\]
This proves \eqref{eq:weighted-discrepancy}.
\end{proof}

Let
\begin{equation}
\theta_t^*
:=
(\Phi^\top D\Phi)^{-1}
\Phi^\top D\mathcal T_\chi\widehat Q_t.
\label{eq:ideal-block}
\end{equation}

\begin{proposition}[One-block parameter error]
\label{prop:block-error}
Under Assumptions~\ref{ass:mixing} and~\ref{ass:features} and the
Q-stage conditions in \eqref{eq:q-stage-stepsize}, every outer block
\(t\) satisfies, almost surely,
\begin{align}
&\E\left[
\norm{\theta_{t,K}-\theta_t^*}_2
\,\middle|\,\mathcal B_t
\right] \quad\le
\varepsilon_Q(K,\ell)
\nonumber\\
&
+\frac{\gamma}{\mu_\phi}
\left[
\varepsilon_{\rm var}(N,\ell)
+\mathcal E_\eta
+\frac{c_\delta}{2}
\left(
\mathcal E_{\rm sc}
+\frac{\mathcal E_{\rm sc}^2}{\ell}
\right)
+\frac{c_\delta\ell}{2}
\right].
\label{eq:block-error}
\end{align}

If \(\alpha_N\) is chosen as in \eqref{eq:main-stepsizes}, then the
same bound holds with \(\varepsilon_{\rm var}(N,\ell)\) replaced by
\(\overline\varepsilon_{\rm var}(N,\ell)\).
\end{proposition}

\begin{proof}
Let
\[
V_t:=V_{\widehat Q_t},
\qquad
\theta_t^{\rm var}
:=
\theta^{\rm var}_{V_t,\bar\nu_t,\bar\Theta_t},
\]
where \((\bar\nu_t,\bar\Theta_t)\) is the Stage~1 output in outer
block \(t\). This output and \(\theta_t^{\rm var}\) are
\(\mathcal H_{t,0}\)-measurable, and
\[
\mathcal B_t=\mathcal F_{b_t}
\subseteq
\mathcal F_{q_t}=\mathcal H_{t,0}.
\]
Therefore, Theorem~\ref{thm:q-stage} gives, almost surely,
\[
\E\left[
\norm{\theta_{t,K}-\theta_t^{\rm var}}_2
\,\middle|\,\mathcal H_{t,0}
\right]
\le
\varepsilon_Q(K,\ell).
\]
Taking conditional expectation with respect to \(\mathcal B_t\) and
using the tower property yields
\begin{equation}
\E\left[
\norm{\theta_{t,K}-\theta_t^{\rm var}}_2
\,\middle|\,\mathcal B_t
\right]
\le
\varepsilon_Q(K,\ell).
\label{eq:q-block-conditional}
\end{equation}

By the definitions of \(\theta_t^{\rm var}\) and \(\theta_t^*\),
\begin{align}
\theta_t^{\rm var}-\theta_t^*
&=
\gamma(\Phi^\top D\Phi)^{-1}
\nonumber\\[-1mm]
&\quad\times
\sum_i d_i\phi(i)
\left[
f_{V_t,\ell}(i;\bar\nu_t,\bar\Theta_t)
-\sigma_i^\chi(V_t)
\right].
\label{eq:var-ideal-difference}
\end{align}
The feature normalization implies that, for arbitrary scalars
\(\{a_i\}\),
\(
\norm{\sum_i d_i a_i\phi(i)}_2
\le
\sum_i d_i|a_i|.
\)
Moreover, Assumption~\ref{ass:features} gives
\(
\norm{(\Phi^\top D\Phi)^{-1}}_{\rm op}
\le
\frac{1}{\mu_\phi}.
\)
Applying Lemma~\ref{lem:target-discrepancy} to
\eqref{eq:var-ideal-difference} therefore yields
\begin{align}
\norm{\theta_t^{\rm var}-\theta_t^*}_2
\le
\frac{\gamma}{\mu_\phi}
\Bigg[
&J_{V_t,\ell}^{\mathrm{PSD},*}
-J_{V_t,\ell}(\bar\nu_t,\bar\Theta_t)
+\mathcal E_\eta
\nonumber\\
&+\frac{c_\delta}{2}
\left(
\mathcal E_{\rm sc}
+\frac{\mathcal E_{\rm sc}^2}{\ell}
\right)
+\frac{c_\delta\ell}{2}
\Bigg].
\label{eq:var-to-ideal-block}
\end{align}
Taking conditional expectation given \(\mathcal B_t\) and applying
Theorem~\ref{thm:stage1} gives
\begin{align}
&\E\left[
\norm{\theta_t^{\rm var}-\theta_t^*}_2
\,\middle|\,\mathcal B_t
\right]
\nonumber\\
&\quad\le
\frac{\gamma}{\mu_\phi}
\left[
\varepsilon_{\rm var}(N,\ell)
+\mathcal E_\eta
+\frac{c_\delta}{2}
\left(
\mathcal E_{\rm sc}
+\frac{\mathcal E_{\rm sc}^2}{\ell}
\right)
+\frac{c_\delta\ell}{2}
\right].
\label{eq:variational-block-error}
\end{align}
Combining \eqref{eq:q-block-conditional} and
\eqref{eq:variational-block-error} using the triangle inequality
proves \eqref{eq:block-error}. Finally,
\eqref{eq:eps-var-simplified} gives the stated bound with
\(\overline\varepsilon_{\rm var}(N,\ell)\).
\end{proof}
\subsection{Across-block contraction and proof of
Theorem~\ref{thm:main}}

\begin{lemma}[Outer target-network recursion]
\label{lem:outer-contraction}
For every deterministic \(\widehat Q_0\in\cQ\) and every integer
\(T\geq0\),
\begin{align}
\E\norm{\widehat Q_T-Q_\chi^*}_\infty
&\leq
\gamma^T\norm{\widehat Q_0-Q_\chi^*}_\infty
\nonumber\\
&\quad+
\sum_{t=0}^{T-1}\gamma^{T-1-t}
\E\norm{\theta_{t,K}-\theta_t^*}_2
+\Gamma_T\mathcal E_Q.
\label{eq:outer-bound}
\end{align}
\end{lemma}

\begin{proof}
This is the robust counterpart of the outer-loop argument in
\cite[Prop.~4.2 and App.~B.1]{chen2023target}.
By the algorithm and the definition of $\theta_t^*$,
\[
\widehat Q_{t+1}
=
\clip(\Phi\theta_{t,K}),
\qquad
\clip(\Pi_\phi\mathcal T_\chi\widehat Q_t)
=
\clip(\Phi\theta_t^*).
\]
Componentwise clipping is nonexpansive, and
$\norm{\phi(i)}_2\leq1$. Therefore,
\begin{align}
\left\|
\widehat Q_{t+1}
-\clip(\Pi_\phi\mathcal T_\chi\widehat Q_t)
\right\|_\infty
&\leq
\max_i
\left|
\phi(i)^\top(\theta_{t,K}-\theta_t^*)
\right|
\nonumber\\
&\leq
\norm{\theta_{t,K}-\theta_t^*}_2.
\label{eq:outer-estimation-step}
\end{align}
Since $\widehat Q_t\in\cQ$, the definition
\eqref{eq:Q-approx} gives
\[
\left\|
\clip(\Pi_\phi\mathcal T_\chi\widehat Q_t)
-\mathcal T_\chi\widehat Q_t
\right\|_\infty
\leq\mathcal E_Q.
\]
Finally, $Q_\chi^*=\mathcal T_\chi Q_\chi^*$, and the exact robust
Bellman operator is a $\gamma$-contraction in $\ell_\infty$. The
triangle inequality therefore gives
\begin{equation}
\norm{\widehat Q_{t+1}-Q_\chi^*}_\infty
\leq
\norm{\theta_{t,K}-\theta_t^*}_2
+\mathcal E_Q
+\gamma\norm{\widehat Q_t-Q_\chi^*}_\infty.
\label{eq:outer-one-step}
\end{equation}
Taking expectations and iterating \eqref{eq:outer-one-step} proves
\eqref{eq:outer-bound}.
\end{proof}

\begin{proof}[Proof of Theorem~\ref{thm:main}]
Taking expectations in Proposition~\ref{prop:block-error} and using
$\varepsilon_{\rm var}(N,\ell)
\leq\overline\varepsilon_{\rm var}(N,\ell)$ gives, uniformly in $t$,
\begin{align*}
\E\norm{\theta_{t,K}-\theta_t^*}_2
\leq{}&
\varepsilon_Q(K,\ell)
+\frac{\gamma}{\mu_\phi}
\Bigg[
\overline\varepsilon_{\rm var}(N,\ell)
+\mathcal E_\eta\\
&\qquad+
\frac{c_\delta}{2}
\left(
\mathcal E_{\rm sc}
+\frac{\mathcal E_{\rm sc}^2}{\ell}
\right)
+\frac{c_\delta\ell}{2}
\Bigg].
\end{align*}
Substitution into \eqref{eq:outer-bound} and
$\sum_{t=0}^{T-1}\gamma^{T-1-t}=\Gamma_T$ give
\eqref{eq:main-bound}.
\end{proof}

\section{Implementation and Representation Tradeoffs}
\label{sec:computation}

The PSD scale class enforces $u(i)\geq\ell$ while preserving affine
dependence on a convex parameter. When entrywise nonnegative features
are available, it can be replaced by
\begin{align}
u_{\rho,\ell}(i)
&=\ell+\zeta(i)^\top\rho,
&
\mathcal M_\rho
&=\{\rho\geq0:\norm{\rho}_2\leq R_\rho\},
\label{eq:two-scale-classes}
\end{align}
where \(\zeta(i)\in\mathbb R_+^{d_\rho}\) and
\(\norm{\zeta(i)}_2\leq1\).
Joint concavity and the Stage~1 proof remain unchanged after replacing
$(\Theta,X_i,R_\Theta)$ by
$(\rho,\zeta(i),R_\rho)$ and modifying the corresponding gradient
bound and scale-approximation error.

The vector model requires $O(d_\rho)$ projection time and storage. In
contrast, a dense PSD parameter requires $O(d_\chi^2)$ storage and an
$O(d_\chi^3)$ eigendecomposition in the worst case. With
$\zeta_j(i)=\chi_j(i)^2$, the vector class is the diagonal subclass of
the PSD class; the latter is more expressive through its cross terms,
whereas the former is computationally cheaper.

For neural approximation, one can replace the linear dual and scale
models in \eqref{eq:sample-objective} by heads
$\eta_\nu(i)$ and $u_\omega(i)=\ell+g_\omega(i)$, with
$g_\omega(i)\geq0$. Their frozen or averaged outputs define the same
one-transition variational targets for Q-regression. Softplus,
sigmoid-scaled, or squared outputs can enforce positivity. This
retains the objective-level reformulation but yields a nonconvex
optimization problem outside the present finite-time analysis.

\section{Experiments}
\label{sec:experiments}

We evaluate two aspects of the proposed method. First, a tabular
MiniCliff experiment compares the learned Q-function with a
high-accuracy robust-DP solution in a setting covered by the theory.
Second, a continuous-state Pendulum experiment tests whether the
variational target can be integrated with neural function approximation
and standard Double DQN components \cite{vanhasselt2016deep}. Code,
frozen configurations, seed lists, and figure-regeneration scripts
are available at
\url{https://github.com/Saptarsh/DRRL-implementations-chi-square}.

\subsection{Tabular experiment}
\begin{figure}[t]
    \centering
    \includegraphics[width=\linewidth]{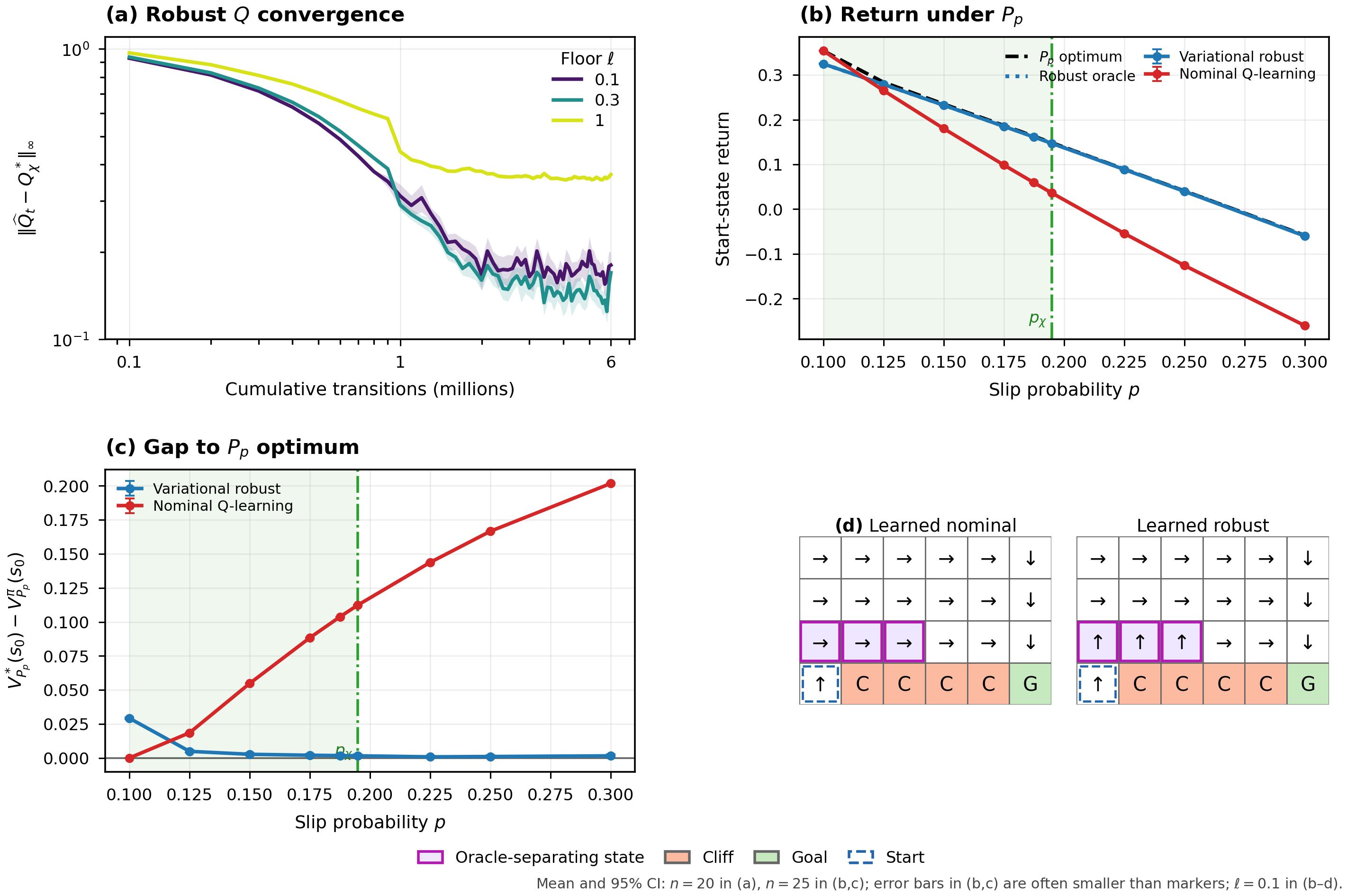}
\caption{MiniCliff results. Panel~(a) plots
\(\|\widehat Q_t-Q_\chi^*\|_\infty
=\max_{s,a}|\widehat Q_t(s,a)-Q_\chi^*(s,a)|\)
against cumulative transitions from the persistent nominal trajectory.
We compute the high-accuracy reference \(Q_\chi^*\) by robust value
iteration on the known MDP: each finite-support row problem is solved by an
active-set calculation to floating-point precision, and iteration stops when
\(\|Q_{k+1}-Q_k\|_\infty\leq10^{-11}\). Moderate floors
\(\ell\in\{0.1,0.3\}\) yield smaller final errors than \(\ell=1\), consistent
with the floor-bias term in the analysis. Panels~(b,c) evaluate fixed policies
under \(P_p\), without Monte Carlo error, by solving
\(V_{P_p}^{\pi}=(I-\gamma P_p^\pi)^{-1}r^\pi\).
The nominal learner outperforms the robust learner near
\(P_{p_0}\), whereas the latter more closely tracks the robust oracle over
the adverse ambiguity-set interval; the \(P_p\)-optimal curve serves as an upper
reference. Panel~(d) shows the corresponding
cliff-avoiding policy change. Thus, the experiment illustrates decreasing
error toward the robust reference and the intended qualitative change in
control. Normal-approximation \(95\%\) intervals 
\(\bar x\pm1.96s/\sqrt n\) use \(n=20\) in panel~(a) and \(n=25\) for the
learned policies in panels~(b,c). The background shading in panels~(b,c)
marks \(p\in[p_0,p_\chi]\), not statistical uncertainty; panel~(d) shows
25-seed modal policies.}
    \label{fig:tabular-results}
\end{figure}

We use a continuing variant of the canonical Cliff Walking problem
\cite[Example~6.6]{sutton1998reinforcement}. Variants of this problem are
also commonly used to visualize risk-sensitive and distributionally robust
policies \cite{liang2024single}. The start state is the lower-left cell (see Figure~\ref{fig:tabular-results}), the
four intermediate bottom-row cells are cliff states, and the lower-right cell
is the goal. Ordinary, cliff, and goal rewards are $-0.01$, $-1$, and $+1$,
respectively; cliff and goal marker states reset to the start on the following
transition. Under \(P_p\), a commanded action is executed with probability \(1-p\),
while each other action is executed with probability \(p/3\). We use $\gamma=0.9$,
$p_0=0.10$, and $\delta=0.10$.

For an interior state--action pair, the four executed actions have distinct
successors, and this row attains the maximum divergence:
\(
\max_{s,a}
D_{\chi}\!\left(
P_p(\cdot\mid s,a)\,\middle\|\,P_{p_0}(\cdot\mid s,a)
\right)
=
\frac{(p-p_0)^2}{p_0(1-p_0)}.
\)
Hence the adverse ambiguity-set interval is
\(
p\in[p_0,p_\chi],\quad
p_\chi=p_0+\sqrt{\delta p_0(1-p_0)}
       =0.194868\ldots .
\)

Each run uses one-hot features and one persistent nominal trajectory
generated by
\[
\pi_b
=
0.8\,\operatorname{Unif}(\mathcal A)
+0.2\,\pi_{\rm goal},
\]
where $\pi_{\rm goal}$ follows the deterministic safe route: up from
the start, right above the cliff, and down in the final column. Thus,
the preferred action has probability $0.4$, every other action has
probability $0.2$, and the observations are Markovian rather than
i.i.d.  Each of 60 blocks reads
$5\times10^4$ transitions to fit $(\eta,u)$ and $5\times10^4$ transitions
to update robust Q directly. Nominal Q-learning updates on every transition
in the same stream, so panel~(a)'s horizontal axis counts total environment
transitions and reaches $6\times10^6$.
\subsection{Neural function approximation}
\label{sec:neural_experiments}

This experiment examines whether the variational target can be combined
with standard deep-Q-learning components in a continuous-state control
problem. It is an implementation study rather than a comprehensive
benchmark against specialized robust-RL algorithms, and its evolving
neural representations lie outside the finite-time theory.

\paragraph{Environment and uncertainty model.}
Pendulum swing-up is a classical nonlinear-control problem 
\cite{aastrom2000swinging} and a long-standing reinforcement-learning
benchmark \cite{doya2000reinforcement}. We use a NumPy implementation of the Gym Pendulum environment
\cite{brockman2016openaigym}, with a discretized torque set and
an additional stochastic actuator-reversal mechanism.

The environment represents a rigid pendulum of mass $m_{\rm pend}=1$ and
length $L_{\rm pend}=1$ rotating about a fixed pivot under gravity $g=10$.
Its latent state is continuous and two-dimensional,
\(
    s=(\theta,\dot\theta),
\)
where $\theta=0$ is the upright position. To avoid the discontinuity at
$\theta=\pm\pi$, the controller observes
\(
    o=(\cos\theta,\sin\theta,\dot\theta).
\)
Thus, the task has a continuous state space rather than a finite number of
states. The controller chooses one of three commanded torques,
\(
    \mathcal{A}=\{-2,0,2\}.
\)

Under \(P_p\), each nonzero commanded torque is independently reversed
with probability \(p\), while the zero action is unchanged. Training uses the nominal kernel \(p_0=0.10\) and $\chi^2$ radius
\(\delta=0.25\). For rows whose commanded and reversed torques have
distinct successors,
\(
D_\chi\!\left(
P_p(\cdot\mid s,a)\,\middle\|\,P_{p_0}(\cdot\mid s,a)
\right)
=
\frac{(p-p_0)^2}{p_0(1-p_0)};
\)
the zero-action row is unchanged and has divergence zero. Hence this one-parameter family
lies in the ambiguity set precisely for \(p\in[0,0.25]\); larger
probabilities are out-of-set stress tests.

Episodes last 200 steps. Training uses discount factor $\gamma=0.99$ and
$0.01$ times the standard Pendulum reward. The plotted raw episodic return is
the undiscounted 200-step sum of the corresponding unscaled rewards; higher
return therefore means lower accumulated control cost.

\begin{figure}[t]
    \centering
    \includegraphics[width=\linewidth]{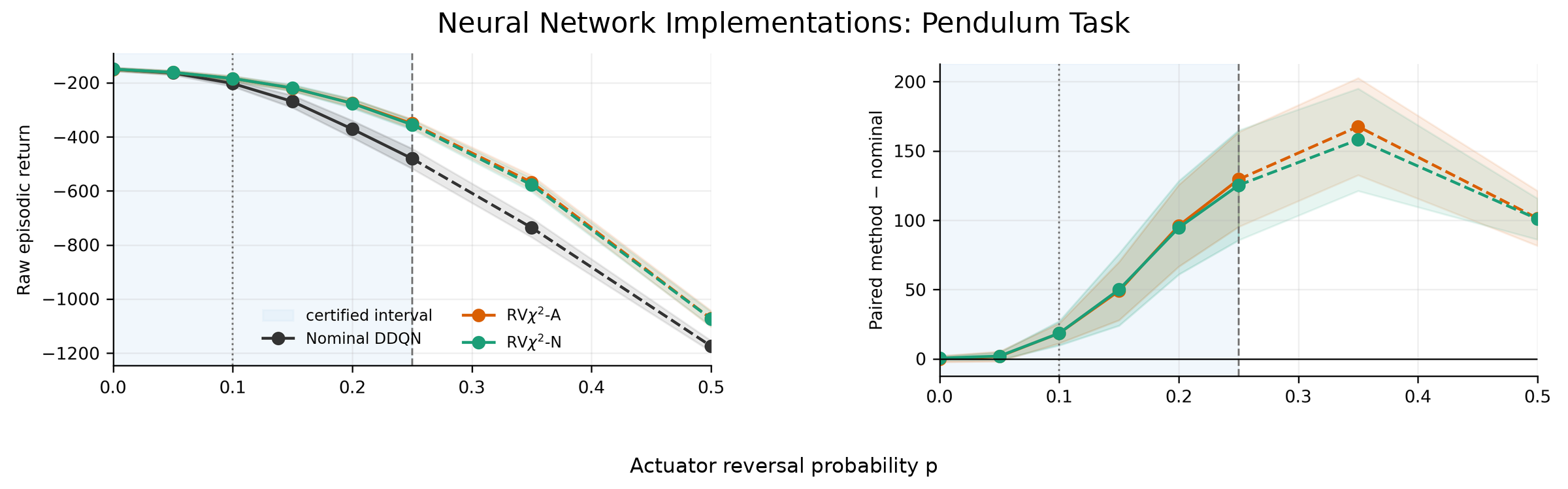}
  \caption{Robustness of the final Pendulum policies under i.i.d.\ actuator
reversals. DDQN denotes Double DQN \cite{vanhasselt2016deep}.
Left: mean raw episodic return. Right: the matched seed-level
robust-minus-nominal return difference; positive values favor the robust
method. Each seed-level value first averages 100 common-random-number
evaluation episodes. Points and bands are means and two-sided pointwise $95\%$
Student-$t$ intervals over the ten reporting-seed observations. The dotted
line marks $p_0=0.10$, the shaded region marks the tested ambiguity-set kernels,
and the dashed line marks their upper boundary $p=0.25$; larger probabilities
are out-of-set stress tests.}   \label{fig:pendulum_nn}
\end{figure}

\paragraph{Neural parameterizations.}
All three methods in Figure~\ref{fig:pendulum_nn} use the same
action-value architecture. The observation $o$ passes through two
fully connected hidden layers with 128 $\tanh$ units each, followed by
a linear output layer with one weight vector $W_a$ for each of the
three actions. Let
\(
    h_\omega(o)\in[-1,1]^m,\quad m=128,
\)
denote the output of the second hidden layer. The Q-function can then be
written as
\[
    Q_{\omega,W}(o,a)
    =
    W_a^\top
    \frac{(1,h_\omega(o))}{\sqrt{m+1}}.
\]
We denote the robust-variational affine and neural auxiliary variants
by RV\(\chi^2\)--A and RV\(\chi^2\)--N, respectively; all methods share the
same Q-network architecture.

RV$\chi^2$--A uses affine dual and scale functions on top of the representation
produced by a frozen target Q-network. At outer block $t$, let
$h_{\bar\omega_t}$ denote this frozen representation, and define
\[
    \psi_t(o)
    =
    \frac{(1,h_{\bar\omega_t}(o))}{\sqrt{m+1}},
    \qquad
    \zeta_t(o)
    =
    \frac{
      \bigl(1,[h_{\bar\omega_t}(o)]_+,
               [-h_{\bar\omega_t}(o)]_+\bigr)
    }{\sqrt{m+1}}.
\]
For each action $a$, the dual and scale functions are
\begin{align}
    \eta_t(o,a)
        = \nu_a^\top\psi_t(o), \quad
    u_t(o,a)
        = \ell+\rho_a^\top\zeta_t(o).
\end{align}
Here $\nu_a\in\mathbb{R}^{m+1}$ and
$\rho_a\in\mathbb{R}^{2m+1}$. The rows of $\nu$ and $\rho$ are projected
onto fixed norm balls, and $\rho_a$ is projected onto the nonnegative
orthant, ensuring that $u_t(o,a)\geq\ell>0$.

The Q encoder is frozen only within the
current auxiliary stage; it is updated later during Q-regression and a new
frozen representation is formed at the next outer block.

RV$\chi^2$--N instead uses two independent multilayer perceptrons applied
directly to
\(
\widetilde o=(\cos\theta,\sin\theta,\dot\theta/8):
\)
\(
F_\eta,F_u:\mathbb{R}^3\longrightarrow\mathbb{R}^3.
\)
Each network has architecture
\(
    3
    \longrightarrow 128\;(\tanh)
    \longrightarrow 128\;(\tanh)
    \longrightarrow 3.
\)
Their outputs are transformed as
\[
    \eta(\widetilde o,\cdot)
    =\eta_{\max}\tanh\!\left(F_\eta(\widetilde o)\right),
\qquad
u(\widetilde o,\cdot)
    =\ell+\operatorname{softplus}\!\left(F_u(\widetilde o)\right).
\]
where $\eta_{\max}=18$ and $\ell=10^{-3}$ in the reported experiment. 

\paragraph{Training protocol.}

The neural implementation preserves the high-level ideas of Algorithm~\ref{alg:main}, but adapts each part to
deep Q-learning. Each simulation (seed) first produces a common 150,000-step nominal
Double-DQN prefix. The resulting Q-network, optimizer state, and replay buffer
are copied to the nominal, RV$\chi^2$--A, and RV$\chi^2$--N branches. Thus,
the branches begin from the same learned controller and the same initial
data. After branching, however, each method follows its own
$\epsilon$-greedy policy and therefore accumulates a method-specific replay
buffer from the same nominal data-generating kernel.

Training then proceeds for 30 outer blocks. In each block, every branch first
collects 3,000 additional nominal-kernel transitions. The current Q-network
of that branch is then copied and held fixed for the remainder of the block.

For each robust branch, replay minibatches fit $\eta$ and $u$ against
successor values from the block-frozen Q-network. During the 5,000 auxiliary
updates, a persistent exponentially averaged copy of the auxiliary parameters
is updated after every optimizer step with decay $0.99$. This copy is then
fixed while the online Q-network receives 1,000 Huber-regression updates using
the corresponding variational Bellman targets.

The nominal branch receives
the same number of Q updates but uses the ordinary Double-DQN target. At the
next block, a new Q snapshot is formed and the process is repeated. All Q and
auxiliary updates use replay minibatches of size 256.

Replay buffers, Adam optimization,
Huber regression, warm starts, exponential moving average deployment, and evolving neural features
also distinguish this implementation from the theorem-covered algorithm.

The Q-learning rate is \(2.5\times10^{-4}\) during nominal pretraining and
\(10^{-4}\) afterward. Hyperparameters were selected using five development
seeds and then frozen before training and evaluation on ten disjoint reporting
seeds; all reported means and intervals use only the latter. Each trained
policy was evaluated for 100 episodes, with initial states and actuator-fault
draws shared across methods within each seed.

\paragraph{Results.}
Figure~\ref{fig:pendulum_nn} shows small paired differences at
$p=0$ and $p=0.05$, followed by increasing robust advantages from the
nominal probability toward the ambiguity-set boundary. At $p=0.25$,
RV$\chi^2$--A improves over its matched nominal branch by
\(
129.44\;[95.27,163.62]
\)
raw-return points, while RV$\chi^2$--N improves it by
\(
125.27\;[85.63,164.91].
\)
The brackets are two-sided pointwise $95\%$ Student-$t$ intervals over
the ten seed-level paired differences.

The two robust parameterizations produce similar mean robustness
profiles, although we did not conduct a superiority test between them.
RV$\chi^2$--A demonstrates that the variational target can be combined
with a frozen learned representation, while RV$\chi^2$--N shows that
the auxiliary functions themselves can be fully neural.

\section{Limitations}

The finite-time guarantee applies to finite MDPs with deterministic
bounded rewards, fixed linear function classes, a positive-definite
stationary Q-feature Gram matrix, a geometrically mixing behavior chain
with full state--action support, compact parameter sets, and exact
projections. For fixed function classes and fixed \(\ell>0\), the bound retains the
approximation and floor terms displayed in
Theorem~\ref{thm:main}. The neural experiment demonstrates practical
implementability of the variational target, but learned representations,
replay-based optimization, and nonconvex auxiliary networks are not
covered by the theory. Extending the analysis to these settings remains
an open problem.

\section{Conclusion}

We studied model-free distributionally robust Q-learning with
state--action-rectangular $\chi^2$ uncertainty, fixed linear function
approximation, and observations from one continuing Markovian
trajectory. An objective-level variational reformulation removes the square root
of a conditional second moment before sampling, yielding conditionally
unbiased one-transition gradients for jointly learned dual and scale
functions. A positive scale floor controls the inverse-scale terms at
the cost of an explicit bias, while blockwise frozen targets separate
within-block estimation from contraction of the exact robust Bellman
operator. This separation
gives a finite-time supremum-norm guarantee for every
$\gamma\in(0,1)$, with distinct statistical, function-approximation,
and floor terms, and a sufficient trajectory length of
$\widetilde O(\epsilon^{-6})$ whenever the function classes support
the prescribed accuracy. The tabular and Pendulum experiments
illustrate, respectively, decreasing error toward a high-accuracy robust-DP reference and the practical use of the variational target with neural
function approximation.

\section*{References}
\bibliographystyle{IEEEtran}
\bibliography{references_new}

\end{document}